\documentclass{article}
\usepackage[english]{babel}

\usepackage[letterpaper,top=2cm,bottom=2cm,left=3cm,right=3cm,marginparwidth=1.75cm]{geometry}

\usepackage{amsmath}
\usepackage[pdftex]{graphicx}
\usepackage[colorlinks=true, allcolors=blue]{hyperref}

\newcommand{\mytitle}{Augmented Risk Prediction for the Onset of Alzheimer's Disease from Electronic Health Records with Large Language Models}

\usepackage{microtype}
\usepackage{geometry}
\usepackage{booktabs} 
\usepackage{bbm}
\usepackage{mathtools}
\usepackage{nccmath}
\usepackage{setspace}

\usepackage{subcaption}

\usepackage{algorithm} 

\usepackage{algpseudocode}

\usepackage[T1]{fontenc}
\usepackage{wrapfig,lipsum,booktabs}

\usepackage[numbers]{natbib}
\usepackage{soul}
\usepackage{dsfont}
\usepackage{enumerate}
\usepackage{enumitem}

\usepackage{amsmath}
\usepackage{amsfonts}
\usepackage{bbm}
\usepackage{dsfont}
\usepackage[Symbol]{upgreek}
\usepackage{lscape}
\usepackage{caption}
\usepackage{balance}
\usepackage{xspace}
\usepackage{float}

\usepackage{soul}

\usepackage{wasysym}
\usepackage{multirow}
\usepackage{array, boldline, rotating}

\usepackage{pifont}

\newcommand{\mc}[1]{\mathcal{#1}}

\renewcommand*\eqref[1]{(\ref{#1})}

\newcommand{\eg}{\emph{e.g.,~}}
\newcommand{\ie}{\emph{i.e.,~}}

\newcommand{\myparagraph}[1]{\vspace{0.07cm}\noindent\textbf{#1}~}

\def\code#1{\texttt{#1}}

\newcommand{\thickhline}{\hlineB{4}}

\usepackage{framed}
\usepackage{xcolor}

\definecolor{LightGray}{gray}{0.9}
\definecolor{ForestGreen}{rgb}{0.0, 0.7, 0.3}
\definecolor{goldenrod}{rgb}{1.0,0.84,0.3}
\definecolor{shadecolor}{named}{LightGray}
\usepackage{tablefootnote}

\usepackage{framed}
\definecolor{shadecolor}{named}{LightGray}

\NewDocumentCommand{\supptitle}{s}{
\onecolumn
\begin{center}
    \rule{\textwidth}{0.03cm}\\[0.1cm]
    -Supplementary Material-\\[0.2cm]
    {\Large 
        \textbf{\mytitle }
    }\\
    \rule{\textwidth}{0.03cm}\\[0.2cm]
\end{center}
}

\newcommand{\refcolor}[2]{
\begingroup\hypersetup{linkcolor=#1}\kern-0.5em#2\endgroup}%

\usepackage{amsmath,amsfonts,bm}

\def\eqref#1{equation~\ref{#1}}

\def\1{\bm{1}}

\DeclareMathAlphabet{\mathsfit}{\encodingdefault}{\sfdefault}{m}{sl}
\SetMathAlphabet{\mathsfit}{bold}{\encodingdefault}{\sfdefault}{bx}{n}

\DeclareMathOperator*{\argmax}{arg\,max}

\hypersetup{
  colorlinks   = true, 
  urlcolor     = black, 
  linkcolor    = ForestGreen, 
  citecolor   = cyan 
}
\begin{document}

\title{\mytitle}

\author{Jiankun Wang$^1$, Sumyeong Ahn$^1$, Taykhoom Dalal$^2$, Xiaodan Zhang$^1$, Weishen Pan$^2$, \\
Qiannan Zhang$^2$, Bin Chen$^1$, Hiroko H. Dodge$^3$, Fei Wang$^2$, and Jiayu Zhou$^1$}
\date{    $^1$Michigan State University, USA \\ 
        $^2$Weill Cornell Medicine, Cornell University, USA \\ 
        $^3$Massachusetts General Hospital, Harvard Medical School, USA 
} 

\providecommand{\keywords}[1]
{
  \small	
  \textbf{\textit{Keywords---}} #1
}

\maketitle
\begin{abstract}
Alzheimer's disease (AD) is the fifth-leading cause of death among Americans aged 65 and older. Screening and early detection of AD and related dementias (ADRD) are critical for timely intervention and for identifying clinical trial participants. The widespread adoption of electronic health records (EHRs) offers an important resource for developing ADRD screening tools such as machine learning based predictive models. 
Recent advancements in large language models (LLMs) demonstrate their unprecedented capability of encoding knowledge and performing reasoning, which offers them strong potential for enhancing risk prediction. 
This paper proposes a novel pipeline that augments risk prediction by leveraging the few-shot inference power of LLMs to make predictions on cases where traditional supervised learning methods (SLs) may not excel. 
Specifically, we develop a collaborative pipeline that combines SLs and LLMs via a confidence-driven decision-making mechanism, leveraging the strengths of SLs in clear-cut cases and LLMs in more complex scenarios. 
We evaluate this pipeline using a real-world EHR data warehouse from Oregon Health \& Science University (OHSU) Hospital, encompassing EHRs from over 2.5 million patients and more than 20 million patient encounters. 
Our results show that our proposed approach effectively combines the power of SLs and LLMs, offering significant improvements in predictive performance. 
This advancement holds promise for revolutionizing ADRD screening and early detection practices, with potential implications for better strategies of patient management and thus improving healthcare.
\end{abstract}


\section{Introduction}
\label{sec:intro}

Alzheimer's disease (AD) and Alzheimer's disease related dementias (ADRD) are neurodegenerative disorders primarily affecting memory and cognitive functions. They gradually erode overall function abilities, eventually leading to death~\cite{mattson2004pathways}. The development of AD/ADRD treatment has been slow due to the complex disease pathology and clinical manifestations. The decline of memory and cognitive functions is associated with pathological progression and structural changes of the brain~\cite{jack2010hypothetical}, which can be identified from neuroimage or biomarkers from cerebro-spinal fluid. However, those procedures are expensive and invasive, which are unlikely to be ordered for asymptomatic patients. For real world patients, typically only the electronic health records (EHRs) collected from their routined care are available\cite{beam2018big, goldstein2017opportunities}. These data include information like demographics, lab tests, diagnoses, medications, and procedures, and they provide a potential opportunity for risk prediction of AD/ADRD~\cite{li2023early}.

Risk prediction from EHRs is commonly formulated as a supervised learning problem~\cite{tang2024leveraging} and one can model with existing supervised learning (SLs) tools, such as logistic regression (LR)~\cite{wu2010prediction}, XGBoost (XGB)~\cite{nori2019machine}, and multi-layer perceptron (MLP)~\cite{shickel2017deep}. However, SL approaches face significant challenges in predicting risk from EHRs, due to the complexity of medical problems and the noisy nature of the data~\cite{zhou2014micro}. Moreover, EHRs do not contain all critical information that is needed for risk prediction for particular conditions. For example, diagnosis of MCI requires a comprehensive evaluation of cognitive functions, such as memory, executive function, and language. In early stages, when symptoms are subtle and not extensively documented in the EHRs, risk prediction using traditional machine-learning approaches can be difficult. Though some information in EHRs may be weakly related to the risk, SL models may or may not be able to pick them up.

Recent advancements in pre-trained large language models (LLMs)~\cite{touvron2023llama, touvron2023llama2, achiam2023gpt, brown2020language, team2023gemini} have demonstrated their capability to provide robust reasoning power, particularly with rich contextual information and domain knowledge. Intuitively, LLM can leverage its reasoning capability and flexible in-context learning (ICL) strategies to better derive valuable insights from EHRs. However, there are still several technical challenges to achieve this goal. The first one is how to perform effective reasoning with an EHR database. While fine-tuning external knowledge into the LLMs has been a major approach in many domains, it is not trivial to fine-tune knowledge from EHR to LLMs. EHR includes clinical information for individual patients and evolves over time, whereas LLMs are typically learned and tuned using static information. The second challenge is the representation of medical records for reasoning. LLMs are probability models trained to understand and reason with natural language, and it is not clear how structured EHRs, such as vital, diagnosis codes, and prescriptions, are best represented in LLMs for effective reasoning. The third challenge is rooted in the inherent data quality issues in EHR data, which could be noisy as they were originally designed for billing purposes. The presence of such events is likely to compromise and greatly mislead the reasoning of LLMs. 

\myparagraph{Contributions.} Here, we summarize the contributions as follows:
\begin{itemize}[leftmargin=10pt]
    \item We identified the strengths and weaknesses of SLs and LLMs in risk predictions from EHR.
    From the SLs' perspective, they provide accurate predictions for confident samples, which are typically aligned well with training data distribution. However, when the samples are not common or the features are sparse, SLs are usually not confident about the predictions and generate poorer predictions than LLMs, showing the value of reasoning from LLMs in EHR analysis.  
    \item Based on our findings, we propose a collaborative approach that combines SLs and LLMs through a confidence-driven selection process for enhanced ADRD risk prediction. This method dynamically selects between SL and LLM predictions based on confidence levels, effectively leveraging the strengths of SLs for high-confidence cases and LLMs for low-confidence instances. Furthermore, we incorporate a meticulously designed ICL demonstration denoising strategy to save the ICL performance of LLMs, which in turn boosts the overall efficiency of the pipeline.
    \item We validate our approach using a real-world dataset from the OHSU health system, highlighting the effectiveness of our method and its superiority over traditional SLs and LLMs in predicting ADRD. Additionally, we conduct experiments with different sizes of LLMs and models fine-tuned on various medical datasets. Our findings suggest that neither a larger model size nor fine-tuning on medical data consistently improves risk prediction performance. Further investigation is required to check these dynamics in practice.
\end{itemize}

\section{OHSU Data and ADRD Risk Prediction}
\label{sec:dataset}
In this section, we describe the real-world dataset that is the primary focus of this paper. It involves a classification task, \ie determining whether a patient has (or will have) an ADRD. We will provide a detailed description of the dataset.

\myparagraph{Dataset overview.}
We utilize extensive EHR data from OHSU Hospital, which is mined by the Oregon Clinical and Translational Research Institute (OCTRI)~\cite{zhang2019metapred}. This dataset contains longitudinal EHRs for $N$ (approx. $N=2.5$ million) individuals (\ie $D = \{(\mathbf{x}_i, y_i)\}_{i=1}^{N}$). Here, $\mathbf{x}_{i}$ represents the feature vector of the $i^\text{th}$ patient and $y_i$ is the corresponding ADRD label, respectively. $\mathbf{x}_i$ is characterized by $d$ features (\ie $\mathbf{x}_i = [x_1,\dots,x_d]$), where $x_d$ denotes the value of the $d^\text{th}$ feature. These $d$ features are categorized into five primary types: vital signs, laboratory test results, International Classification of Diseases (ICD) codes, RxNorm medication codes, and Current Procedural Terminology (CPT) codes. Detail information for each category is described in~\autoref{tab:feature}. This rich information is further processed as features that support the development of risk predictive models for identifying ADRD, along with analysis of different ADRD patient group definitions and prediction windows~\cite{zhou2014micro}. We split $D$ into training $D_{\text{train}}$ and testing $D_{\text{test}}$ datasets after obtaining $D$ by following the procedures described in this section.

\begin{table*}[t]
    \centering
    \resizebox{\textwidth}{!}{%
    \begin{tabular}{>{\centering\arraybackslash}p{1.5cm}|
                    >{\centering\arraybackslash}p{4.6cm}
                    >{\centering\arraybackslash}p{4.6cm}
                    >{\centering\arraybackslash}p{3.8cm}
                    >{\centering\arraybackslash}p{4.2cm}
                    >{\centering\arraybackslash}p{4.4cm}}
        \thickhline
        \textbf{Type}                & \textbf{Vital sign}            &   \textbf{Lab. Test}           &   \textbf{ICD}                         &   \textbf{RxNorm}                      & \textbf{CPT}    \\ \hline
        Domain              & $\mathbb{R}$          &   $\mathbb{R}$            &  [0,1] (Positive, Negative)   &   [0,1] (Positive, Negative)  & [0,1] (Positive, Negative)    \\ \hline
        Example             & Blood Pressure, Age   &   Hemoglobin Level    &  \code{J18.9} for Pneumonia   &   \code{4099} for Estrogen    & \code{A4206} for DME and supplies            \\ \hline
        Short Explanation   
        &   Physiological measurement to assess a patient's status
        &   Analyzing biochemical markers using blood and urine 
        &   Alphanumeric system classifying diseases
        &   Standardized nomenclature system for clinical drugs                            
        &   Medical procedure identification for billing          \\
        \thickhline
    \end{tabular}}
    \caption{Brief explanation of the five categories; Vital sign, Laboratory test, ICD code, RxNorm code, and CPT code, in the EHR dataset, describing each patient.}
    \label{tab:feature}
\end{table*}

\subsection{Task Description}
Following existing risk prediction models~\cite{albrecht2018predicting}, we formulate the risk prediction as classification tasks that distinguish between ``\code{case}'' and ``\code{control}'' labels, which correspond to ``positive'' and ``negative'' outcomes, respectively, in the context of ADRD. The definition of ADRD \code{cases} (labeled as ``positive'') is established through a combined methodology involving the use of ICD-Ninth and -Tenth Revision (ICD-9/ICD-10) diagnosis codes along with the administration of specific anti-dementia medications, namely \emph{donepezil}, \emph{galantamine}, \emph{memantine}, and \emph{rivastigmine}. This classification approach is consistent with existing computable phenotypes documented in the literature, such as those found in references~\cite{li2023early, wilkinson2019identifying}, with the exception of \emph{aducanumab}, which was excluded due to its recent FDA approval in 2021 and the temporal constraints of our dataset. \autoref{table:cp_rules} presents the two rule-based \emph{computable phenotypes} (CPs) employed in this research. These CP algorithms were meticulously designed to accurately identify and categorize the various forms of ADRD, acknowledging the heterogeneity inherent in these conditions. \code{Controls} are defined as those without ADRD-related diagnoses or other conditions potentially associated with or causative of dementia (\eg mild cognitive impairment, Parkinson's disease, and so on.), and who had not been prescribed dementia-related medications.

\begin{table}[t]
\centering
\resizebox{0.8\columnwidth}{!}{%
\begin{tabular}{l|p{6cm}|l|l}
\thickhline
\textbf{Idx} & \textbf{Computable phenotype (CP) rules} & \multicolumn{1}{c|}{\textbf{Disease}} & \multicolumn{1}{c}{\textbf{Citation}} \\ \hline
\multirow{3}{*}{CP1} & \multirow{3}{6cm}{patients have at least one encounter with a relevant ADRD diagnosis code} & ADRD, other dementias\tablefootnote{Includes Alzheimer's disease (AD), vascular dementia, Lewy body dementia, frontotemporal dementia, senile dementia, Parkinson's disease dementia, Huntington's dementia, HIV-related dementia, alcohol-related dementia, and unspecified dementia.} & Wilkinson et al.~\cite{wilkinson2019identifying} \\ \cline{3-4} 
 && ADRD, other dementias\tablefootnote{Includes AD, vascular dementia, frontotemporal dementia, alcohol-related dementia, memory loss, and unspecified dementia.} & Fujiyoshi et al.~\cite{fujiyoshi2017validity}\\ \cline{3-4} 
 && AD & Tjandra et al.~\cite{tjandra2020cohort}  \\ \hline
\multirow{2}{*}{CP2} & \multirow{2}{6cm}{patients have at least two encounters with relevant ADRD diagnosis codes} & ADRD, other dementias\textsuperscript{1} & Wilkinson et al.~\cite{wilkinson2019identifying} \\ \cline{3-4}
 && AD & Wei et al.~\cite{wei2015extracting} \\ \thickhline
 \end{tabular}}\\
\raggedright
\caption{The definition of ADRD-related CPs.}
\label{table:cp_rules}
\end{table}

\begin{figure*}[t]
    \centering
    \includegraphics[width=0.98\textwidth]{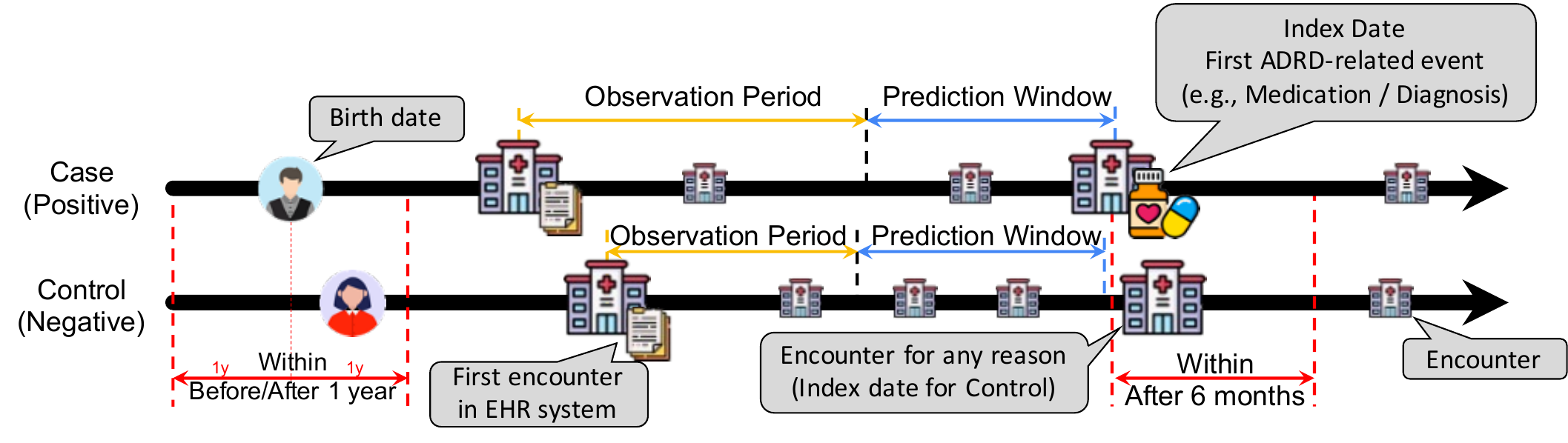}
    \caption{Summary of constructing \code{control} samples for each \code{case}: Among patients, we select a subset that has a time gap of within one year compared to the \code{case}, and also has an encounter within six months after the \code{case} is diagnosed with ADRD. The observation period is the duration between the first encounter date and the start of the prediction window, which is selected from among \{0, 1, 3\} years. In short, a longer prediction window implies predicting a further future possibility of ADRD.}
    \label{fig:time_line}
\end{figure*}

\subsection{Constructing \code{Control} Set}
Three steps are involved in constructing the \code{control} set based on the \code{cases} of each computable phenotype. The first step filters out samples whose birth date and encounter are far from those of \code{cases} (see \textcolor{red}{red lines} in~\autoref{fig:time_line}). After that, we select samples with higher similarity scores to the \code{cases}. Finally, we split the previously obtained dataset into three different prediction window cases for each CP (see \textcolor{blue}{blue lines} in~\autoref{fig:time_line}).

\myparagraph{Select similar birth date and encounter statistics.}
In the first step, we select patients with similar birth dates and encounter statistics. More precisely, we choose patients from the entire \code{control} set (\ie not \code{cases}) who were born within one year before or after the \code{case}, and who have a history of visiting the hospital within six months following the \code{case} diagnosis date (which will be called ``index date''). As described in \textcolor{red}{red lines} of~\autoref{fig:time_line}, the birth date of the \code{control} lies within one year and also includes encounters within six months of the related \code{case}.

\myparagraph{Similar sample mining based on the propensity score.}
As the second step, we select 10 \code{controls} for each \code{case} by utilizing propensity scores, thereby constructing a data category ratio that simulates real-world scenarios. We adhere to the following steps to select the \code{controls} based on the propensity score: (1) The comorbidity score is measured using ICD codes~\cite{li2023early}. (2) Subsequently, the propensity score is computed using LR~\cite{austin2011introduction}. (3) We then construct a distance matrix between the propensity scores of \code{case} and \code{controls}. (4) Finally, we find the optimal matching using the Hungarian algorithm over the distance matrix generated in the third step. Note that after this sample mining process, we have \code{control} and \code{cases} with statistics of Prediction Window 0 (PW0) in~\autoref{tab:config}. Other cases (PW1, PW3) are described in the next paragraph, explaining how we subsample the PW0 case.

\begin{table}[t]
    \centering
    \resizebox{0.5\columnwidth}{!}{%
    \begin{tabular}{c|cc|cc|cc}
        \thickhline
            & \multicolumn{2}{c|}{PW0} & \multicolumn{2}{c|}{PW1} & \multicolumn{2}{c}{PW3} \\ 
            & Control & Case & Control & Case & Control & Case  \\\hline
        CP1 & 26,870 & 2,687 & 25,618 & 2,376 & 23,826 & 1,929 \\
        CP2 & 20,190 & 2,019 & 19,290 & 1,786 & 18,064 & 1,473 \\
        \thickhline
    \end{tabular}}
    \caption{Dataset configuration after sample mining based on propensity scores.}
    \label{tab:config}
\end{table}

\myparagraph{Timeline of \code{case} and \code{control}.}
To conduct \emph{risk} prediction of ADRD, we define the prediction window. We identify the first ADRD-related diagnosis or medication prescription date for \code{cases}, \ie positive, termed the ``index date.'' The corresponding \code{controls}, \ie negatives, which are selected from the second step, regard the encounter dates within six months after the ``index date'' of the case as the ``index date'' for them. The timeline prior to the index date is divided into a prediction window (in \{0, 1, 3\} years before the index date)  and an observation period (from the first database encounter to the start of the prediction window, see \textcolor{goldenrod}{yellow lines} in~\autoref{fig:time_line}). Predictive analyses utilized data solely from the observation period to explore the feasibility of ADRD prediction at various lead times, with a minimum one-year observation period required for inclusion. 

\subsection{Additional Processing}
\label{sec 2.3}
\myparagraph{Outlier handling.}
Since real-world datasets usually contain missing values or human-error-induced incorrect values, we employ the $3$-sigma rule~\cite{pukelsheim1994three} as a systematic method for outlier detection. By applying this criterion, we can identify extreme values that might lead to misleading predictions. We then set these outliers directly to \code{NaN}, in preparation for imputation. Given the inevitable presence of missing values in the dataset, especially within laboratory results and vital signs, we suggest imputing them using the median value of each feature from the training data. This method is robust to outliers and preserves the central tendency of the data distribution, making it appropriate for medical scenarios.

\myparagraph{Decoding features to reduce sparsity.}
To address the challenge of processing sparsely encoded values such as ICD codes in our dataset, we implement a decoding strategy that enhances accessibility following previous work~\cite{li2023early}. Specifically, the 18,922 unique ICD-9/10 codes are converted into 1,712 Phenotype codes, such as transforming \code{J18.9} (\emph{Pneumonia, unspecified organism}) into \code{480} (\emph{Pneumonia}). Similarly, we simplify $25,281$ RxNorm medication codes to $909$ ingredient codes and $2,332$ CPT codes to $40$ Clinical Classification Software codes, significantly reducing data sparsity. It is important to note that the preprocessed dataset is in table format. For each patient group, defined by their computable phenotype and prediction window, we exclude any laboratory tests with more than 50\% missing data and applied one-hot encoding to binary features. We then create a feature vector for each patient, which consists of continuous laboratory tests and vital signs data, combined with one-hot encoded vectors for PheCodes, RxNorm ingredient codes, and CCS codes. Despite the above efforts, the resulting feature vectors are still relatively sparse.

\section{Preliminary and Motivation}
\label{sec:motivation}
In this section, we present the motivation behind our proposed method, highlighting the intuitive pros and cons of SLs and LLMs in the analysis of EHRs. We then offer empirical evidence to validate our insights and illustrate the potential of augmenting risk prediction with LLMs using in-context learning (ICL). Before describing our motivation, we briefly summarize how ICL works.

\subsection{Preliminary of ICL}
Suppose that we have the training dataset $D_{\text{train}} = \{(\mathbf{x}_i, y_i)\}_{i=1}^{n}$ where input text is $\mathbf{x}_i$ and its corresponding output text is $y_i$. For any test query $\mathbf{x}$\footnote{In this study, $\mathbf{x}$ refers to tabular data or text generated from tabular data. This simplification facilitates the processing of the same data $\mathbf{x}$ by both SLs and LLMs.}, we get LLM's response $\hat{y}_{\text{\scriptsize LLM}}$ using the following ICL paradigm:
\begin{align*}
    \hat{y}_{\text{\scriptsize LLM}} =  \ \code{LLM}(\mc{T}(D_{\text{ICL}}^{\mathbf{x}}, \mathbf{x})), \quad \text{where} \quad \mc{T}(D_{\text{ICL}}^{\mathbf{x}}, \mathbf{x}) =  ({\mathbf{x}}_1, {y}_1) \circ ({\mathbf{x}}_2, {y}_2) \circ \ldots \circ ({\mathbf{x}}_k, {y}_k) \circ \mathbf{x},
\end{align*}
and $\{({\mathbf{x}}_j, {y}_j)\}_{j=1}^{k} = D_{\text{ICL}}^{\mathbf{x}} \subset D_{\text{train}}$ is the selected samples for $\mathbf{x}$ from the training dataset and $\circ$ denotes the concatenation operation. To construct the ${D}_{\text{ICL}}^{\mathbf{x}}$, there are several approaches, such as a term-frequency-based approach or utilizing neural networks to select the informatively relevant samples. Here, ICL is a way of achieving better inference using LLMs by feeding relevant examples to the model so that these examples serve as hints for understanding the context. From these relevant examples provided, LLMs are expected to make better reasoning in answering the question. It is one of the most cost-efficient methods for adapting to the target task since we can make LLMs adapt to our context \emph{without fine-tuning}.

\subsection{Pros and Cons of SLs and LLMs using ICL}
To create a synergistic combination of SLs and LLMs, we describe the pros and cons of each SL and ICL hereinafter.

\begin{table*}[t!]
     \begin{minipage}[b]{0.49\textwidth}
     \centering
        \resizebox{0.9\columnwidth}{!}{
        \begin{tabular}{c|c|c|c|c}
            \thickhline
            \multicolumn{1}{c|}{Dataset} & \multicolumn{1}{c|}{Model}  & \multicolumn{1}{c|}{Precision} & \multicolumn{1}{c|}{Recall} & \multicolumn{1}{c}{F1 score}  \\ 
            \hline
            \multirow{5}{*}{CP1 / PW1} 
            & LR 
            & .168 \footnotesize{$\pm .027$}      & .616 \footnotesize{$\pm .088$}      & .262 \footnotesize{$\pm .034$}  \\
            \cline{2-5}
            & XGB 
            & .233 \footnotesize{$\pm .018$}      & .392 \footnotesize{$\pm .035$}      & .292 \footnotesize{$\pm .024$}  \\
            \cline{2-5}
            & MLP 
            & .209 \footnotesize{$\pm .037$}      & .443 \footnotesize{$\pm .032$}      & .283 \footnotesize{$\pm .041$}  \\
            \cline{2-5}
            & Avg. SLs 
            & .230 \footnotesize{$\pm .028$}      & .444 \footnotesize{$\pm .029$}      & \textcolor{blue}{\textbf{.302} \footnotesize{$\pm .029$}}  \\
            \cline{2-5}
            & LLaMA2 70B 
            & .150 \footnotesize{$\pm .009$}      & .394 \footnotesize{$\pm .045$}      & .217 \footnotesize{$\pm .015$}  \\
            \cline{1-5}
            \multirow{5}{*}{CP2 / PW1} 
            & LR 
            & .230 \footnotesize{$\pm .009$}      & .615 \footnotesize{$\pm .050$}      & .335 \footnotesize{$\pm .016$}  \\
            \cline{2-5}
            & XGB 
            & .264 \footnotesize{$\pm .030$}      & .500 \footnotesize{$\pm .060$}      & .345 \footnotesize{$\pm .038$}  \\
            \cline{2-5}
            & MLP 
            & .248 \footnotesize{$\pm .014$}      & .541 \footnotesize{$\pm .078$}      & .339 \footnotesize{$\pm .027$}  \\
            \cline{2-5}
            & Avg. SLs 
            & .263 \footnotesize{$\pm .014$}      & .536 \footnotesize{$\pm .075$}      & \textcolor{blue}{\textbf{.352} \footnotesize{$\pm .028$}}  \\
            \cline{2-5}
            & LLaMA2 70B 
            & .151 \footnotesize{$\pm .019$}      & .417 \footnotesize{$\pm .046$}      & .222 \footnotesize{$\pm .027$}  \\
            \thickhline
        \end{tabular}
        }
        \captionof{table}{Performance of \textcolor{blue}{easy (confident)} samples}
        \label{tab:conf_perf}
     \end{minipage}
     \hfill
     \begin{minipage}[b]{0.49\textwidth}
         \centering
         \resizebox{0.9\columnwidth}{!}{
        \begin{tabular}{c|c|c|c|c}
            \thickhline
            \multicolumn{1}{c|}{Dataset} & \multicolumn{1}{c|}{Model} & \multicolumn{1}{c|}{Precision} & \multicolumn{1}{c|}{Recall} & \multicolumn{1}{c}{F1 score}  \\ 
            \hline
            \multirow{5}{*}{CP1 / PW1} 
            & LR 
            & .110 \footnotesize{$\pm .026$}      & .822 \footnotesize{$\pm .100$}      & .192 \footnotesize{$\pm .038$}  \\
            \cline{2-5}
            & XGB 
            & .148 \footnotesize{$\pm .054$}      & .233 \footnotesize{$\pm .127$}      & .178 \footnotesize{$\pm .077$}  \\
            \cline{2-5}
            & MLP 
            & .072 \footnotesize{$\pm .037$}      & .487 \footnotesize{$\pm .271$}      & .125 \footnotesize{$\pm .065$}  \\
            \cline{2-5}
            & Avg. SLs 
            & .109 \footnotesize{$\pm .038$}      & .519 \footnotesize{$\pm .217$}      & .180 \footnotesize{$\pm .064$}  \\
            \cline{2-5}
            & LLaMA2 70B 
            & .160 \footnotesize{$\pm .030$}      & .542 \footnotesize{$\pm .070$}      & \textcolor{red}{\textbf{.247} \footnotesize{$\pm .041$}}  \\
            \cline{1-5}
            \multirow{5}{*}{CP2 / PW1} 
            & LR 
            & .141 \footnotesize{$\pm .039$}      & .868 \footnotesize{$\pm .014$}      & .240 \footnotesize{$\pm .059$}  \\
            \cline{2-5}
            & XGB 
            & .120 \footnotesize{$\pm .070$}      & .204 \footnotesize{$\pm .129$}      & .148 \footnotesize{$\pm .087$}  \\
            \cline{2-5}
            & MLP 
            & .123 \footnotesize{$\pm .057$}      & .613 \footnotesize{$\pm .219$}      & .202 \footnotesize{$\pm .089$}  \\
            \cline{2-5}
            & Avg. SLs 
            & .148 \footnotesize{$\pm .068$}      & .551 \footnotesize{$\pm .200$}      & .232 \footnotesize{$\pm .102$}  \\
            \cline{2-5}
            & LLaMA2 70B 
            & .188 \footnotesize{$\pm .050$}      & .546 \footnotesize{$\pm .062$}      & \textcolor{red}{\textbf{.273} \footnotesize{$\pm .049$}}  \\
            \thickhline
        \end{tabular}
        }
         \captionof{table}{Performance of \textcolor{red}{hard (unconf.)} samples}
         \label{tab:unconf_perf}
     \end{minipage}
\end{table*}

\myparagraph{Pros and cons of SLs.} 
The traditional SLs, such as LR, XGB, and MLP, are notable primarily for their strong learning capability to extract common patterns from data. For example, MLP trains the model parameters to extract meaningful features among various input features by iteratively feeding and backpropagating the training dataset multiple times. As illustrated in~\refcolor{blue}{\autoref{tab:conf_perf}}, SLs perform better than LLMs on \emph{easy (confident) samples}—those samples that closely align with the common patterns recognized during training. These samples are explicitly marked by their high levels of predictive confidence. Note that merging the knowledge of SLs (Average SLs) yields better F1 scores than when a single SL is used, demonstrating greater robustness.

However, there are several challenges with SLs. One significant limitation is their difficulty in handling \emph{hard (unconfident) samples}—those with sparse or uncommon features compared to the training data. SLs tend to generate predictions with low confidence for these challenging cases. As illustrated in~\refcolor{red}{\autoref{tab:unconf_perf}}, the inference quality on hard samples is notably poorer compared to that of LLMs. This issue arises because SLs struggle to extract valuable clues from samples that diverge significantly from common patterns.

\myparagraph{Pros and cons of LLMs using ICL.}
On the other hand, LLMs exhibit distinct characteristics over SLs. The main strength of LLMs is their strong pre-trained reasoning capability, which is extensively studied in the research field of ICL. Their reasoning power enables them to effectively manage \emph{hard (unconfident) samples}, particularly when provided with appropriate ICL examples that supplement the hard samples with specific domain knowledge for better prediction. This superiority is evident in~\refcolor{red}{\autoref{tab:unconf_perf}}, where LLMs using ICL demonstrate better performance in addressing hard samples compared to SLs.

However, despite this easy adaptation to each test sample, there are various drawbacks to using LLMs with ICL. As described in~\refcolor{blue}{\autoref{tab:conf_perf}}, it is difficult to capture the common patterns for prediction compared to using SLs, thus we get worse performance on \emph{easy (confident) samples}. This is because ICL only uses a very small number of samples as context from the training dataset. Therefore, it is generally inappropriate to apply LLMs directly to infer each sample in our task, and it is better to use them in a smart manner.

\myparagraph{Integrating SLs and LLMs for ADRD Risk Prediction.} 
Considering the advantages and disadvantages of SLs and LLMs, we are motivated to integrate the strengths of these two methodologies while mitigating their weaknesses for risk prediction. Our insights consist of two key characteristics: (1) When SL generates a confident prediction on a sample, we will use the SL prediction. These samples are typically not hard samples and have rich features, and therefore LLM is unlikely to generate good performance over SL here. (2) When SL generates a low-confident prediction, this suggests a hard sample and usually has low feature quality, and we will use LLMs w/ICL to augment the prediction. We expect that the LLMs can use their reasoning power to make improved predictions on these samples. Based on our intuition, we describe our method in~\autoref{sec:method}.

\begin{figure*}[t]
    \centering
    \includegraphics[width=1.\textwidth]{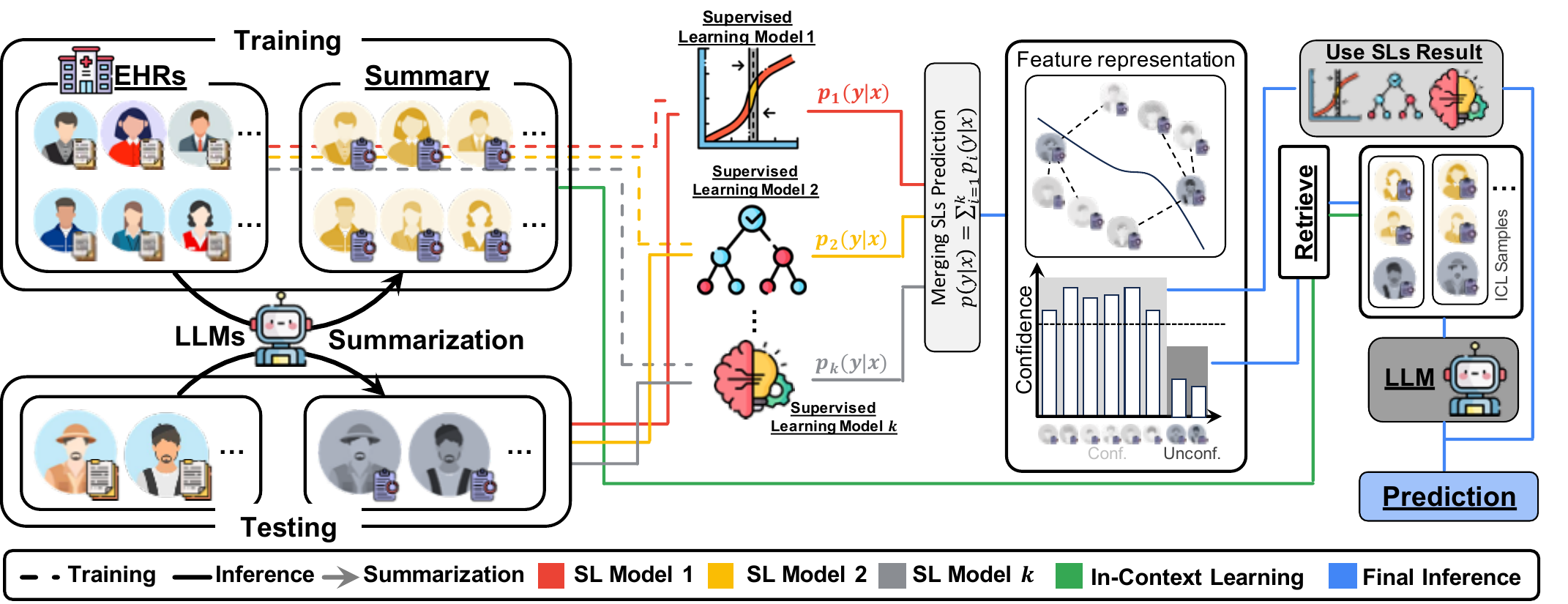}
    \caption{Framework Overview. We combine the power of supervised learning methods and advanced Large Language Models (LLMs) to build risk prediction for the onset of AD/ADRD from Electronic Health Records (EHRs). We first use LLM to summarize the tabular data in EHRs. Then, we train SLs (\eg LR, XGB, MLP), and split the confidently/unconfidently predicted samples. Finally, we perform in-context sample selection for the unconfident samples and predict them using LLMs.}
    \label{fig:method}
\end{figure*}
\section{Method}
\label{sec:method}
Based on the insights above, we describe our proposed algorithm for predicting ADRD risk, which fuses the advantages of LLMs and SLs. Our approach is tailored such that LLMs target hard samples, utilizing their advanced reasoning capabilities, while SLs efficiently predict easy samples by capitalizing on patterns learned during training. We present the proposed method below, with a brief illustration provided in~\autoref{fig:method}.

\subsection{Risk Prediction on EHRs using ICL}
\myparagraph{Natural Language Summarization of EHRs.}
The EHR structured data is tabular (e.g., ~\cite{zhou2014micro}) but not natural language that is easily understandable by LLMs. For LLMs to consume the information in the structured data, we propose first summarizing the tabular data into natural language using LLMs' summarization capability. Specifically, for a given tabular representation of a patient, the goal is to summarize the information into a comprehensive representation in natural language. LLM is capable of such transformation, and yet there is a trade-off between the details to be included in the summary that is best for downstream in-context learning and risk prediction: While looking for an informative summary, LLM may suffer from hallucination for too excessive elaboration on the summary. We describe the LLM-based summarization of EHRs in~\autoref{app:summary}.

\myparagraph{Reliable subset sampling.}
The next question is how to build $D_{\text{ICL}}^{\mathbf{x}}$ for each test sample. Existing studies (e.g., \cite{bolucu2023impact, liu2022makes}) suggest that the choice of ICL examples will significantly impact performance. Furthermore, the quality of ICL examples is also a key factor that affects ICL performance~\cite{wei2023symbol, li2023finding}. In our framework, we first construct a reliable candidate set for quality ICL examples using SL signals, as we believe the predictability of an ICL example is directly related to the quality of the example. We set the criteria for reliable candidates as ``High Average Prediction Confidence,'' which means we select the samples from the training set whose average prediction probability, across all SLs, is above a predefined threshold, \ie
\begin{equation*}
    D_{\text{reliable}} = \left\{ (\mathbf{x}_i, y_i) \in D_{\text{train}} \, \middle| \, \frac{1}{|\mathcal{A}|} \sum_{A \in \mathcal{A}} p_{\text{\scriptsize A}}({y}_i|\mathbf{x}_i) \geq \tau \right\},
\end{equation*}
where \( \mathcal{A} \) is the set of SL methods, \ie \{LR, XGB, MLP\} in our case, \( \mathbf{x}_i \) is the $i$-th sample in the \emph{training set}, \( y_i \) is the corresponding ground-truth label, and \( p_{A}({y}_i|\mathbf{x}_i) \) represents the predicted probability\footnote{See supplement for details on obtaining the predicted probability for training samples.} by model \( A \). The threshold \( \tau \) serves as a level of confidence that determines whether a sample's prediction is reliable. The defined set \( D_{\text{reliable}} \) contains samples exceeding \( \tau \) in average prediction confidence, deemed robust and quality candidates from the \emph{training set} for building $D_{\text{ICL}}^{\mathbf{x}}$.

\myparagraph{Retrieving score of EHRs for ICL.} 
To construct $D_{\text{ICL}}^{\mathbf{x}}$ for each test sample using reliable candidate set $D_{\text{reliable}}$ , we measure the similarity between test sample $\mathbf{x}$ and candidate sample $\mathbf{x}_j$, where $(\mathbf{x}_j, y_j) \in D_{\text{reliable}}$, based on two types of similarity scores. As the features in vital signs and lab tests have continuous values while the others are categorical features, we divide them into two cases and measure the similarity using Euclidean distance for continuous features, and Hamming distance for categorical features, respectively. Then, we measure the total similarity score $s(\mathbf{x},\mathbf{x}_{j})$ as follows:
\begin{equation*}
s(\mathbf{x},\mathbf{x}_{j}) = \lambda \times d_{\text{Euclidean}}(\mathbf{x}, \mathbf{x}_{j}) + (1 - \lambda) \times d_{\text{Hamming}}(\mathbf{x}, \mathbf{x}_{j}),
\end{equation*}
where $d_{\text{Euclidean}}(\mathbf{x}, \mathbf{x}_{j})$ and $d_{\text{Hamming}}(\mathbf{x}, \mathbf{x}_{j})$ represent the Euclidean and Hamming distances between $\mathbf{x}$ and $\mathbf{x}_j$ using continuous and binary values, respectively. Moreover, $\lambda$ denotes the balancing hyperparameter between these two distance metrics. We select the top-$k$ similar samples for each $\mathbf{x}$ and assign the order of constructing $\mc{T}(D_{\text{ICL}}^{\mathbf{x}}, \mathbf{x})$ in an ascending manner, \ie $(\mathbf{x}_1, y_1)$ and $(\mathbf{x}_k, y_k)$ have the lowest and highest similarity scores among the top-$k$ samples, respectively, so that the model encounters the most relevant samples right before inference~\cite{zhao2021calibrate}.

We note that many approaches for patient similarity retrieval, such as~\cite{zhu2016measuring, hermanutz1975clinical, wang2015psf}, have been developed in the past decade, and they can be naturally adopted in our proposed framework. 
A comparison of existing patient similarity approaches under our framework is important and will be conducted in future work. 

\subsection{Confidence-Driven Prediction}
\myparagraph{Confidence-driven selection.}
A crucial step in our approach involves routing a risk prediction case to either SLs or LLMs. We employ an average confidence score from SL methods as the primary selection criteria, which is defined as follows:
\begin{equation*}
    p(\hat{y}|\mathbf{x})=\frac{1}{|\mathcal{A}|} \sum_{A \in \mathcal{A}} p_{\text{\scriptsize A}}(\hat{y}|\mathbf{x}),
\end{equation*}
where $\mathbf{x}$ is the \emph{test} sample, $\hat{y}$ refers to any possible label in the current task, and $p_{\text{\scriptsize A}}(\hat{y}|\mathbf{x})$ represents the prediction probability from the model A trained on the entire \emph{training set}. According to the confidence score $p(\hat{y}|\mathbf{x})$, we regard the sample as confidently predicted by SLs when the sample has greater than $\sigma$ confidence, \ie $\exists\ \hat{y}\text{, }p(\hat{y}|\mathbf{x}) \ge \sigma$. 
Otherwise, we regard it as an unconfident sample.

\myparagraph{SLs for confident samples.}
For the samples which has higher confidence when we use SLs combination, it has to be answered by using SLs prediction itself based on our motivation and its empirical verification. Therefore, we get the answer $\hat{y}_{\text{\scriptsize SL}}$ of the test sample $\mathbf{x}$ by using the following way:
\begin{equation*}
    \hat{y}_{\text{\scriptsize SL}} = \argmax_{\hat{y}} p(\hat{y}|\mathbf{x}),
\end{equation*}
where $p(\hat{y}|\mathbf{x}) \ge \sigma$. Note that $\sigma$ is the hyperparameter balancing the usages of SLs and LLMs. An increase in $\sigma$ will result in more use of LLMs, while a decrease in $\sigma$ will enhance the reliance on SLs.

\myparagraph{ICL for unconfident samples.}
On the other side, when the test samples cannot be confidently predicted by SLs, \ie $\forall \hat{y}\text{, }p(\hat{y}|\mathbf{x}) < \sigma$, we use ICL to predict them by leveraging the relevant information:
\begin{equation*}
    \hat{y}_{\text{\scriptsize LLM}} = \code{LLM}(\mc{T}(D_{\text{ICL}}^{\mathbf{x}}, \mathbf{x})), 
\end{equation*}
where $\mc{T}(D_{\text{ICL}}^{\mathbf{x}}, \mathbf{x})$ is the concatenated input, and $D_{\text{ICL}}^{\mathbf{x}}$ represents the in-context examples obtained from the reliable candidate set $D_{\text{reliable}}$. The information from $D_{\text{ICL}}^{\mathbf{x}}$ is often more useful than the main pattern from the full training set, as discussed in~\autoref{sec:motivation}, which guides LLM to answer more accurately in low-confidence part.

\subsection{Risk Prediction combining SLs and LLMs}
Based on the above components, we summarize our pipeline to the risk prediction in Algorithm~\autoref{alg:propose}. The proposed algorithm is composed of two phases: \emph{preparation} and \emph{inference}. In the \emph{preparation} phase,  we first train the SLs using the entire training dataset $D_{\text{train}}$ and get the prediction probability $p_{\text{\scriptsize A}}(\hat{y}|\mathbf{x})$ of the test sample $\mathbf{x}$, then compute average confidence score using the signals from all SLs. In the \emph{inference} phase, we first decide whether the test sample $\mathbf{x}$ is confidently answered by SLs or not. If it is confident, we directly use the prediction from SLs. Otherwise, we use the prediction from LLM via ICL. Finally, our system outputs the predicted answer $\hat{y}$. 

\begin{algorithm}[tb]
    \caption{Proposed Algorithm}\label{alg:propose}
    \begin{algorithmic}
        \State \textbf{Input: } SLs (denotes as $\mathcal{A}$), \code{LLM}, $D_{\text{train}}$, $D_{\text{reliable}}$, $\sigma$
        \vspace{2mm}
        \State \textbf{/* Preparation */}
        \For{$A \in $ $\mathcal{A}$}
            \State Train SL model $A$ using $D_{\text{train}}$
            \State \text{Obtain }$p_{\text{\scriptsize A}}(\hat{y}|\mathbf{x})$
        \EndFor
        \State $p(\hat{y}|\mathbf{x})=\frac{1}{|\mathcal{A}|} \sum_{A \in \mathcal{A}} p_{\text{\scriptsize A}}(\hat{y}|\mathbf{x})$ \Comment{Get Avg. confidence score}
        \vspace{3mm}
        \State \textbf{/* Inference when the test input is $\mathbf{x}$ */}
        \If{$\exists\ \hat{y}\text{, }p(\hat{y}|\mathbf{x}) \ge \sigma$} \Comment{Predict via SL}
            \State $\hat{y}_{\text{\scriptsize SL}}  = \argmax_{\hat{y}} p(\hat{y}|\mathbf{x})$
            \State \textbf{Return: } $\hat{y}_{\text{\scriptsize SL}}$
        \Else \Comment{Predict via LLM}
            \State Extract $D_{\text{ICL}}^{\mathbf{x}} \subset D_{\text{reliable}}$ , and build $\mc{T}(D_{\text{ICL}}^{\mathbf{x}}, \mathbf{x})$
            \State $\hat{y}_{\text{\scriptsize LLM}}  = \code{LLM}(\mc{T}(D_{\text{ICL}}^{\mathbf{x}}, \mathbf{x}))$
            \State \textbf{Return: } $\hat{y}_{\text{\scriptsize LLM}}$
        \EndIf
        \vspace{2mm}
    \end{algorithmic}
\end{algorithm}

\section{Experiment}
\label{sec:exp}
\subsection{Experimental Setting}
\myparagraph{Data Splitting.}
To address potential biases and ensure consistency across various computable phenotypes (CP) and prediction windows (PW) intervals, we adopted a uniform random sampling approach, selecting 5500 samples for each CP\_PW group, and maintaining the \code{case}-\code{control} ratio to be 1:10. This method guarantees that the size of the data remains constant across all groups, thereby ensuring that differences do not compromise the comparability of experimental outcomes. Then, we randomly select 80\%, 20\% of total samples to be training and testing sets respectively.

\begin{table*}
\centering
\resizebox{1.0\textwidth}{!}{%
    \begin{tabular}{cccccccccc}
    \thickhline
         &  &  \multicolumn{4}{c}{CP1} & \multicolumn{4}{c}{CP 2} \\ 
        \cmidrule{4-6} \cmidrule{8-10}
         &  & & Precision &  Recall & F1 Score & & Precision & Recall & F1 Score \\ 
         \hline
         & LR   
         & & .1914 \footnotesize{$\pm .0233$} & .6200 \footnotesize{$\pm .0395$} & .2910 \footnotesize{$\pm .0258$}
         & & .1859 \footnotesize{$\pm .0243$} & .6420 \footnotesize{$\pm .0591$} & .2867 \footnotesize{$\pm .0268$} \\
         & XGB
         & & .2255 \footnotesize{$\pm .0219$} & .3960 \footnotesize{$\pm .0689$} & .2849 \footnotesize{$\pm .0237$}
         & & .2719 \footnotesize{$\pm .0435$} & .5160 \footnotesize{$\pm .0557$} & \textbf{.3559 \footnotesize{$\pm .0304$}} \\
        \parbox[t]{2mm}{\multirow{-1}{*}{\rotatebox[origin=c]{90}{PW0}}} 
         & MLP   
         & & .2044 \footnotesize{$\pm .0198$} & .4660 \footnotesize{$\pm .0582$} & .2826 \footnotesize{$\pm .0201$}
         & & .2189 \footnotesize{$\pm .0162$} & .4640 \footnotesize{$\pm .1183$} & .2938 \footnotesize{$\pm .0266$}\\ 
         & Avg. SLs   
         & & .2233 \footnotesize{$\pm .0297$} & .4700 \footnotesize{$\pm .0486$} & .3015 \footnotesize{$\pm .0310$}
         & & .2420 \footnotesize{$\pm .0098$} & .4760 \footnotesize{$\pm .0933$} & .3182 \footnotesize{$\pm .0210$}\\
         & LLaMA2 70B 
         & & .1573 \footnotesize{$\pm .0128$} & .4100 \footnotesize{$\pm .0447$} & .2271 \footnotesize{$\pm .0189$}
         & & .1633 \footnotesize{$\pm .0212$} & .4560 \footnotesize{$\pm .0786$} & .2403 \footnotesize{$\pm .0336$}\\
         & Ours   
         & & .2494 \footnotesize{$\pm .0268$} & .4860 \footnotesize{$\pm .0427$} & \textbf{.3271 \footnotesize{$\pm .0150$}}
         & & .2609 \footnotesize{$\pm .0162$} & .4980 \footnotesize{$\pm .0801$} & .3407 \footnotesize{$\pm .0257$}\\
         \hline
         & LR   
         & & .1554 \footnotesize{$\pm .0261$} & .6420 \footnotesize{$\pm .0796$} & .2484 \footnotesize{$\pm .0338$}
         & & .1996 \footnotesize{$\pm .0164$} & .6580 \footnotesize{$\pm .0426$} & .3058 \footnotesize{$\pm .0212$} \\
         & XGB
         & & .2246 \footnotesize{$\pm .0109$} & .3720 \footnotesize{$\pm .0319$} & .2799 \footnotesize{$\pm .0171$}
         & & .2443 \footnotesize{$\pm .0281$} & .4520 \footnotesize{$\pm .0397$} & .3171 \footnotesize{$\pm .0329$} \\
        \parbox[t]{2mm}{\multirow{-1}{*}{\rotatebox[origin=c]{90}{PW1}}} 
         & MLP   
         & & .1769 \footnotesize{$\pm .0171$} & .4540 \footnotesize{$\pm .0472$} & .2541 \footnotesize{$\pm .0233$}
         & & .2054 \footnotesize{$\pm .0214$} & .5540 \footnotesize{$\pm .0956$} & .2987 \footnotesize{$\pm .0354$}\\ 
         & Avg. SLs   
         & & .1972 \footnotesize{$\pm .0272$} & .4520 \footnotesize{$\pm .0508$} & .2744 \footnotesize{$\pm .0352$}
         & & .2299 \footnotesize{$\pm .0240$} & .5420 \footnotesize{$\pm .0813$} & .3220 \footnotesize{$\pm .0363$}\\
         & LLaMA2 70B 
         & & .1518 \footnotesize{$\pm .0056$} & .4140 \footnotesize{$\pm .0361$} & .2220 \footnotesize{$\pm .0108$}
         & & .1564 \footnotesize{$\pm .0114$} & .4380 \footnotesize{$\pm .0435$} & .2303 \footnotesize{$\pm .0177$}\\
         & Ours   
         & & .2157 \footnotesize{$\pm .0231$} & .5300 \footnotesize{$\pm .0379$} & \textbf{.3064 \footnotesize{$\pm .0285$}}
         & & .2442 \footnotesize{$\pm .0247$} & .5560 \footnotesize{$\pm .0524$} & \textbf{.3386 \footnotesize{$\pm .0294$}}\\
         \hline
         & LR   
         & & .1754 \footnotesize{$\pm .0137$} & .6280 \footnotesize{$\pm .0479$} & .2737 \footnotesize{$\pm .0181$}
         & & .1478 \footnotesize{$\pm .0760$} & .5460 \footnotesize{$\pm .2753$} & .2319 \footnotesize{$\pm .1180$} \\
         & XGB
         & & .2221 \footnotesize{$\pm .0082$} & .4940 \footnotesize{$\pm .0845$} & .3051 \footnotesize{$\pm .0233$}
         & & .2492 \footnotesize{$\pm .0091$} & .4180 \footnotesize{$\pm .0117$} & .3121 \footnotesize{$\pm .0072$} \\
        \parbox[t]{2mm}{\multirow{-1}{*}{\rotatebox[origin=c]{90}{PW3}}} 
         & MLP   
         & & .1945 \footnotesize{$\pm .0065$} & .5100 \footnotesize{$\pm .0927$} & .2799 \footnotesize{$\pm .0167$}
         & & .1903 \footnotesize{$\pm .0198$} & .5020 \footnotesize{$\pm .0458$} & .2756 \footnotesize{$\pm .0253$}\\ 
         & Avg. SLs   
         & & .2090 \footnotesize{$\pm .0060$} & .5140 \footnotesize{$\pm .0806$} & .2957 \footnotesize{$\pm .0150$}
         & & .2334 \footnotesize{$\pm .0116$} & .4840 \footnotesize{$\pm .0294$} & .3149 \footnotesize{$\pm .0205$}\\
         & LLaMA2 70B 
         & & .1003 \footnotesize{$\pm .0217$} & .2280 \footnotesize{$\pm .0440$} & .1392 \footnotesize{$\pm .0289$}
         & & .1513 \footnotesize{$\pm .0131$} & .3860 \footnotesize{$\pm .0301$} & .2172 \footnotesize{$\pm .0173$}\\
         & Ours   
         & & .2248 \footnotesize{$\pm .0133$} & .5220 \footnotesize{$\pm .0601$} & \textbf{.3136 \footnotesize{$\pm .0199$}}
         & & .2375 \footnotesize{$\pm .0140$} & .5140 \footnotesize{$\pm .0174$} & \textbf{.3247 \footnotesize{$\pm .0144$}}\\
    \thickhline
    \end{tabular}}
    \caption{Experiment results on the OHSU dataset. The best results of each metric are highlighted in boldface.}
    \label{tab:main}
\end{table*}

\myparagraph{SL Methods.}
In our study, we compare our pipeline with three SLs commonly used for medical form data \ie LR~\cite{hosmer2013applied}, XGB~\cite{chen2015xgboost} and MLP~\cite{haykin1998neural}, which are also components in our proposed pipeline. These models serve as baselines for medical tabular data classification due to their scalability, and ability to handle complex relationships in features. We also introduce the \textsc{Avg. SLs} approach, which ensembles the above approaches and makes robust predictions as compared to a single model. To address the class imbalance problem in training SL models (see details in~\autoref{tab:config}), we integrated the Synthetic Minority Over-sampling Technique (SMOTE)~\cite{chawla2002smote}, ensuring a balanced class distribution for training. We use 5-fold cross-validation to obtain the optimal hyperparameters of each model, the grid search spaces are detailed in the supplement.

\myparagraph{LLM Configuration.}
In our pipeline, we utilized LLaMA2 as the backbone language model, an open-source LLM developed by Meta and accessed through Hugging Face\footnote{\url{https://huggingface.co/meta-llama}}. Our approach involves employing two distinct versions of the LLaMA2, specifically the 7B model and 70B model, to cater to different aspects of our task. The 7B model is designated for generating patient summaries, whereas the more powerful 70B model is tasked with the complex process of final reasoning. Given the medical field requires more reliable outputs, we use the ``Greedy'' strategy for each output. Specifically, by setting ``num\_beams=1'' (\ie the number of hypotheses to keep track of) and ``do\_sample=false,'' we prevent LLaMA2 from generating outputs other than the most probable one. This approach reduces diversity but enhances the reliability of the results, ensuring reproducibility. For ICL, we use $10$ demonstrations unless noted otherwise. The parameter $\lambda$, set at $0.05$, weights categorical features (like diagnosis and medication) more heavily than continuous features (such as vital signs) in our demonstration retrieval process.

\subsection{Main Result}
To evaluate the performance of our pipeline, we conducted experiments on the processed OHSU dataset and compared the testing performance with those obtained by using SLs and LLM independently. We focus on the OHSU data for six different combinations of computable phenotypes (CP1, CP2) and prediction windows (PW0, PW1, PW3), see~\autoref{table:cp_rules} for definitions. The confidence threshold $\sigma$ in our pipeline is set to 0.6, \ie if the prediction probability produced by \textsc{Avg. SLs} is greater than 0.6, then we use the predicted label of \textsc{Avg. SLs}. Otherwise, we will take the prediction of ICL, which uses demonstrations from the reliable set $D_{\text{reliable}}$.

We repeat the experiment 5 times and report the mean with standard deviation in precision, recall, and F1 score, by~\autoref{tab:main}. We mainly focus on the F1 score, which is the most important metric in classification tasks with label imbalance problems. From~\autoref{tab:main}, we can see our pipeline shows superiority in F1 score among almost all cases, which shows the effectiveness of our proposed pipeline, combining the strength of traditional SLs and LLMs in risk prediction. Our pipeline is capable of managing imbalanced data to ensure robust and accurate predictions. For the baselines, XGB, however, shows consistently good performance in terms of precision, while low performance in recall, suggesting that it tends to predict more negatives. LR's lower precision and higher recall indicate a tendency to answer positives, leading to many misdiagnoses. \textsc{Avg. SLs} achieves a balance among different SL models. LLaMA2 70B model performs the worst in all cases, which shows that it is difficult to capture the common pattern by relying on just ICL. 

Note that the overall poor performance, with F1 scores below 0.4, may be attributed to inherent challenges in the ADRD risk prediction task. In the OHSU dataset, crucial factors like demographics and clinical notes, which could significantly enhance predictive accuracy, are not available. From the perspective of different CPs, the models' performance for CP2 is better than for CP1 in most cases. This is probably because the patients grouped by CP2 have many ADRD-related characteristics that lead them to be diagnosed with multiple ADRDs.

\subsection{Empirical Analysis}
To better understand the proposed pipeline, we conduct some ablation studies: (1) assessing the influence of LLM-based summarization, (2) investigating the consequences of randomly selecting demonstrations in ICL, (3) exploring the different denoising strategies in ICL, (4) hyperparameter sensitivity, \ie studying on different confidence thresholds in the proposed pipeline, (5) the impact of different types of LLMs, \ie analyzing the impact of model size and medical data fine-tuning on predicting performance.

\begin{table}[h]
\centering
\resizebox{0.6\columnwidth}{!}{%
    \begin{tabular}{ccccccc}
    \thickhline
         &&   & Precision &  Recall & F1 Score \\ \hline
         & w/ Summary
         &&  .1583 \footnotesize{$\pm .0087$} & .5733 \footnotesize{$\pm .0205$} & \textbf{.2481 \footnotesize{$\pm .0122$}} \\
        \parbox[t]{2mm}{\multirow{-2}{*}{\rotatebox[origin=c]{90}{CP1}}} 
         & w/o Summary
         &&  .1029 \footnotesize{$\pm .0037$} & .8867 \footnotesize{$\pm .0189$} & .1844 \footnotesize{$\pm .0063$} \\ \hline
         & w/ Summary
         &&  .1686 \footnotesize{$\pm .0129$} & .5733 \footnotesize{$\pm .0249$} & \textbf{.2605 \footnotesize{$\pm .0179$}} \\
        \parbox[t]{2mm}{\multirow{-2}{*}{\rotatebox[origin=c]{90}{CP2}}} 
         & w/o Summary
         &&  .1010 \footnotesize{$\pm .0023$} & .8300 \footnotesize{$\pm .0141$} & .1801 \footnotesize{$\pm .0038$} \\ 
    \thickhline
    \end{tabular}}
    \caption{Performance comparison of with and without summarization module on PW0}
    \label{tab:summarize}
\end{table}
\myparagraph{Effect of LLM-based Summarization.}
In our study, to make predictions based on the LLM, we have to reformulate the tabular data into text. We proposed to use LLM for patient data summarization, and the summarized text will, to some extent, reflect the characteristics of the LLM's pre-trained knowledge, as the model's behavior and output are based on the data distributions and patterns it learned during the training phase. So, we expect that the generated text will be LLM-friendly, which would be useful for subsequent tasks. To confirm the effect of such LLM-generated text, we conduct experiments to compare the use of concatenated sentence only (w/o Summary) and LLM-based summarization in our case (w/ Summary) on CP1\_PW0 and CP2\_PW0 datasets. Here, we set the number $k$ of ICL examples to be 6 because the length of the concatenated sentence is too long and will exceed the input token limit of LLaMA2 if $k=10$. The experiment result in~\autoref{tab:summarize} shows that the pipeline w/ Summary has better performance, as the f1 score is much higher, which indicates the effectiveness of LLM-based summarization. 
See supplement for examples of concatenated sentences and LLM-generated summaries.

\begin{table}[h]
\centering
\resizebox{0.62\columnwidth}{!}{%
    \begin{tabular}{ccccccc}
    \thickhline
         &&  & Precision &  Recall & F1 Score \\ \hline
         & w/ Sim-Selection
         && .2494 \footnotesize{$\pm .0268$} & .4860 \footnotesize{$\pm .0427$} & \textbf{.3271 \footnotesize{$\pm .0150$}} \\
        \parbox[t]{2mm}{\multirow{-2}{*}{\rotatebox[origin=c]{90}{CP1}}} 
         & w/o Sim-Selection
         &&  .2016 \footnotesize{$\pm .0726$} & .3260 \footnotesize{$\pm .1317$} & .2253 \footnotesize{$\pm .0324$} \\ \hline
         & w/ Sim-Selection
         &&  .2609 \footnotesize{$\pm .0162$} & .4980 \footnotesize{$\pm .0801$} & \textbf{.3407 \footnotesize{$\pm .0257$}} \\
        \parbox[t]{2mm}{\multirow{-2}{*}{\rotatebox[origin=c]{90}{CP2}}} 
         & w/o Sim-Selection
         &&  .1656 \footnotesize{$\pm .0314$} & .3700 \footnotesize{$\pm .2208$} & .1997 \footnotesize{$\pm .0144$} \\ 
    \thickhline
    \end{tabular}}
    \caption{Performance comparison of with and without similarity-based sample selection module on PW0}
    \label{tab:simselection}
\end{table}

\myparagraph{Similarity-based Demonstration Retrieval in ICL.}
In our study, we select the in-context examples based on the similarity between the test sample and the reliable candidate sample. The context samples, which are similar to the test sample, are expected to help LLM learn more about the background to make decisions. To evaluate the effect of such a similarity-based retrieval strategy, we conduct experiments on CP1\_PW0 and CP2\_PW0 datasets using two types of ICL sample selection strategy: random selection (w/o Sim-Selection) and similarity-based sample selection (w/ Sim-Selection). Both strategies selected samples from the reliable candidate set $D_{\text{reliable}}$. ~\autoref{tab:simselection} shows that employing a similarity-based sample selection strategy enhances performance across all metrics, suggesting that providing contextually similar examples yields more accurate responses. In contrast, random selection tends to produce a higher proportion of negative answers, reflecting the inherent label imbalance where a predominance of negative ICL examples results in more negative predictions.

\begin{table}[h]
\centering
\resizebox{0.63\columnwidth}{!}{%
    \begin{tabular}{ccccc}
    \thickhline
         &  &  Precision &  Recall & F1 Score \\ \hline
         & \small Full Development Set
         &  .1573 \footnotesize{$\pm .0128$} & .4100 \footnotesize{$\pm .0447$} & .2271 \footnotesize{$\pm .0189$} \\
        \parbox[t]{2mm}{\multirow{-0.8}{*}{\rotatebox[origin=c]{90}{CP1}}} 
         & \small All-Correct Subset
         &  .1901 \footnotesize{$\pm .0094$} & .5660 \footnotesize{$\pm .0242$} & .2844 \footnotesize{$\pm .0114$} \\ 
         & \small Any-Correct Subset
         &  .1838 \footnotesize{$\pm .0039$} & .5220 \footnotesize{$\pm .0306$} & .2717 \footnotesize{$\pm .0066$} \\ 
         & \small High Confidence Subset
         &  .2080 \footnotesize{$\pm .0169$} & .5400 \footnotesize{$\pm .0525$} & \textbf{.2998 \footnotesize{$\pm .0226$}} \\
         \hline
         & \small Full Development Set
         &  .1633 \footnotesize{$\pm .0212$} & .4560 \footnotesize{$\pm .0786$} & .2403 \footnotesize{$\pm .0336$} \\ 
        \parbox[t]{2mm}{\multirow{-0.8}{*}{\rotatebox[origin=c]{90}{CP2}}} 
         & \small All-Correct Subset
         &  .2036 \footnotesize{$\pm .0113$} & .6100 \footnotesize{$\pm .0316$} & .3052 \footnotesize{$\pm .0152$} \\ 
         & \small Any-Correct Subset
         &  .2067 \footnotesize{$\pm .0119$} & .5640 \footnotesize{$\pm .0524$} & .3023 \footnotesize{$\pm .0195$} \\
         & \small High Confidence Subset
         &  .2194 \footnotesize{$\pm .0143$} & .5740 \footnotesize{$\pm .0578$} & \textbf{.3172 \footnotesize{$\pm .0226$}} \\ 
    \thickhline
    \end{tabular}}
    \caption{Performance of different ICL demonstration denoising strategies on PW0}
    \label{tab:reliable}
\end{table}

\myparagraph{Study on Different Denoising Strategies in ICL.}
In our pipeline, we build the reliable candidate set $D_{\text{reliable}}$ for ICL by using signals from conventional SLs, specifically, LR, XGB, and MLP. We hypothesize that those samples that exhibit high prediction confidence across all SLs are high-quality samples, otherwise, they are considered as noise samples. Such a denoising process is expected to boost ICL performance. We test our hypothesis using the CP1\_PW0 and CP2\_PW0 datasets, comparing four strategies for constructing ICL candidate sets. We avoid confidence-driven predictions in our pipeline to better observe performance differences among ICL strategies. The full training set refers to using the complete training set as the candidate. The \emph{all-correct} subset comprises samples that all SLs successfully predict. The \emph{any-correct} subset includes any samples that at least one SL successfully predicts. The high confidence subset, which is the strategy employed in this paper, considers the average confidence levels of predictions across different SLs. According to the results in~\autoref{tab:reliable}, all three denoising strategies significantly enhanced the performance of ICL, with the High Confidence Subset achieving the best results.

\begin{figure}[h]
    \centering
    \begin{subfigure}[b]{0.3\textwidth}
        \includegraphics[width=\textwidth]{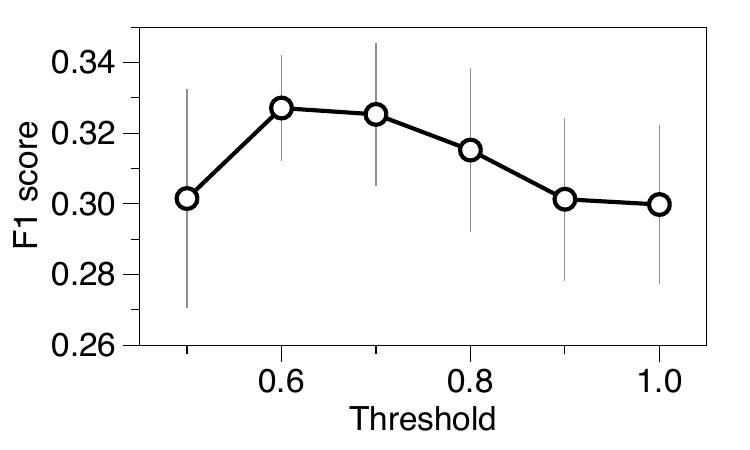}
        \vspace{-15pt}
        \caption{CP1/PW0}
        \label{fig:cp1pw0}
    \end{subfigure}
    \hfill
    \begin{subfigure}[b]{0.3\textwidth}
        \includegraphics[width=\textwidth]{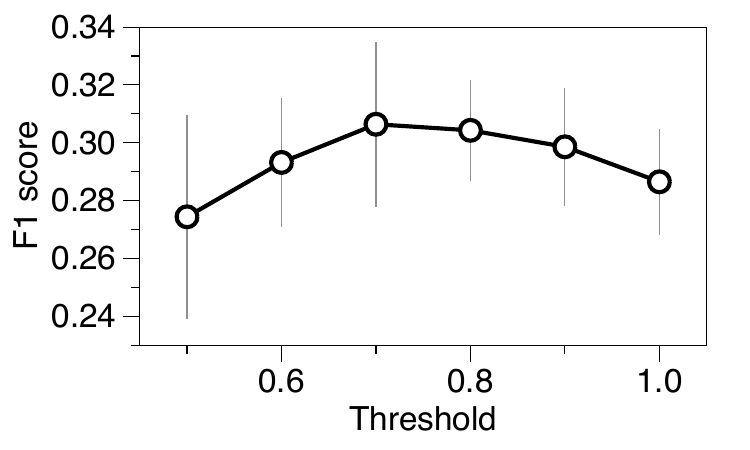}
        \vspace{-15pt}
        \caption{CP1/PW1}
        \label{fig:cp1pw1}
    \end{subfigure}
    \hfill
    \begin{subfigure}[b]{0.3\textwidth}
        \includegraphics[width=\textwidth]{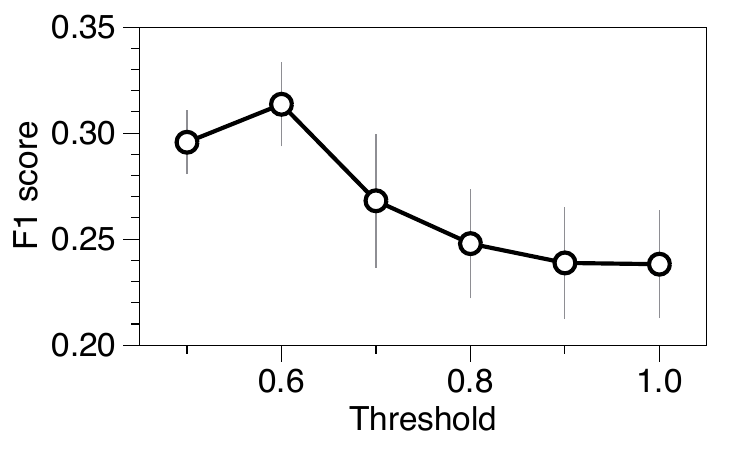}
        \vspace{-15pt}
        \caption{CP1/PW3}
        \label{fig:cp1pw3}
    \end{subfigure}
    
    \begin{subfigure}[b]{0.3\textwidth}
        \includegraphics[width=\textwidth]{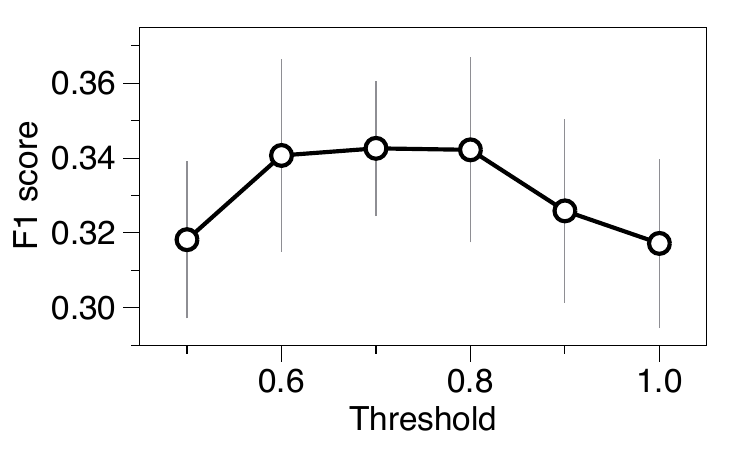}
        \vspace{-15pt}
        \caption{CP2/PW0}
        \label{fig:cp2pw0}
    \end{subfigure}
    \hfill
    \begin{subfigure}[b]{0.3\textwidth}
        \includegraphics[width=\textwidth]{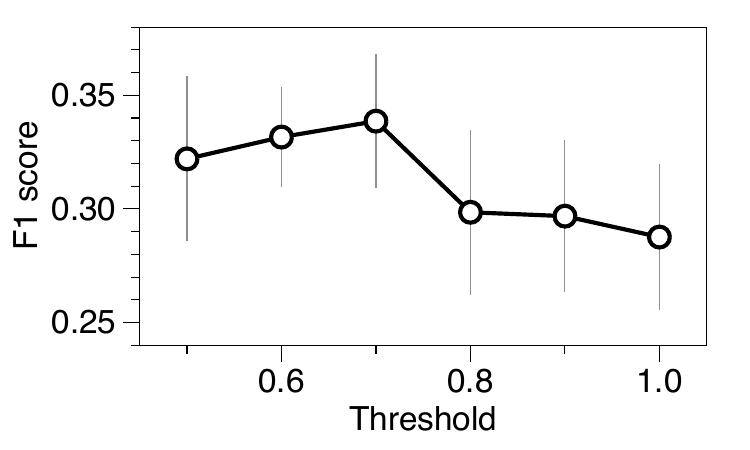}
        \vspace{-15pt}
        \caption{CP2/PW1}
        \label{fig:cp2pw1}
    \end{subfigure}
    \hfill
    \begin{subfigure}[b]{0.3\textwidth}
        \includegraphics[width=\textwidth]{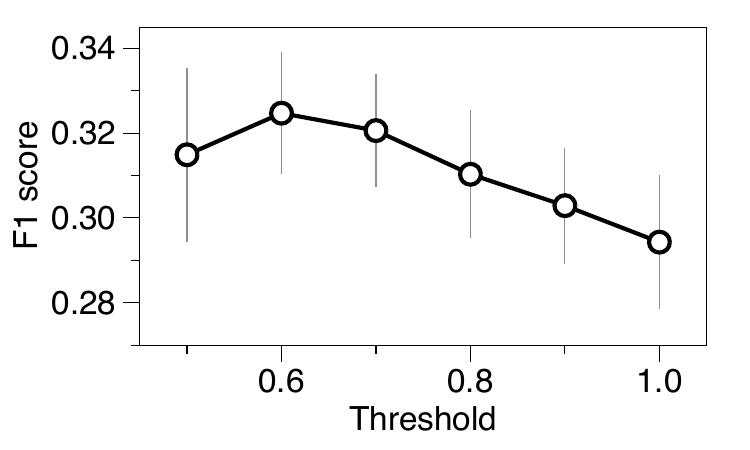}
        \vspace{-15pt}
        \caption{CP2/PW3}
        \label{fig:cp2pw3}
    \end{subfigure}
    \caption{Analysis of different confidence thresholds used in our pipeline on CP\{1, 2\}/PW\{0, 1, 3\}.}
    \label{fig:threshold}
\end{figure}
\myparagraph{Analysis of Different Confidence Thresholds.}
Our pipeline integrates SLs and LLMs for risk prediction through a confidence-driven mechanism. If the prediction probability of \textsc{Avg. SLs} falls below the confidence threshold $\sigma$, the prediction will be generated by the LLM. To investigate the performance at different confidence thresholds, we conduct experiments on CP\{1,2\}\_PW\{0,1,3\} data using six thresholds \{0.5, 0.6, 0.7, 0.8, 0.9, 1.0\}. According to the results in~\autoref{fig:threshold}, we can see that the performance is better with a concave form through confidence threshold, \ie SLs an LLMs balance is necessary. This is consistent with our discussion in~\autoref{sec:motivation}. In other words, LLMs do not outperform SLs on all samples, but confident samples still rely on SLs that capture common patterns via training on extensive data, and unconfident samples are better handled by using LLMs. 

\begin{figure}[ht]
    \centering
    \includegraphics[width=0.7\columnwidth]{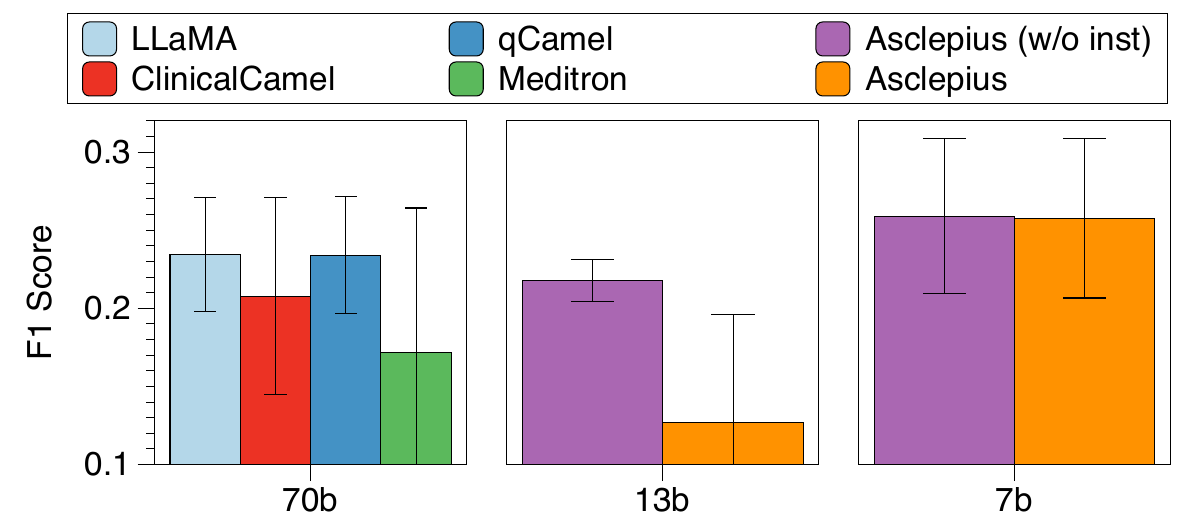}
    \caption{Comparison of performance across different model sizes and fine-tuning on specific medical datasets.}
    \label{fig:diff_llm_chart}
\end{figure}
\myparagraph{The Impact of Model Size and Medical Data Fine-tuning.}
In our approach, we employed the LLaMA2 70B model for inference. To understand the effect of different model sizes on performance, we conducted a series of comparative experiments with various sizes of LLMs. In addition to the pre-trained LLaMA2 models, we also compared models that were fine-tuned on diverse medical datasets to assess their efficacy on our ADRD prediction task. See the supplement for details about the medical data fine-tuned models used in our experiment. The results in~\autoref{fig:diff_llm_chart} indicate that medical data fine-tuning does not necessarily enhance performance for the target task. The effectiveness depends on the similarity between the domain of the data used for fine-tuning and the target data domain. For example, the Asclepius-7b model, which performed the best, utilized a fine-tuning dataset containing a large number of QA pairs related to ADRD. This dataset closely aligns with our target task, which likely contributed to the model's superior performance. Furthermore, our findings challenge the common wisdom from scaling laws that \emph{larger models are inherently better}; we observed that smaller models can achieve superior performance under certain conditions. This highlights the need for a nuanced understanding of the interaction between model size and domain-specific fine-tuning within the context of task-specific applications.
\section{Related Work}
\label{sec:related}
We categorize related research of this work into the following categories: (1) the use of machine learning for handling ADRD, (2) large language models, (3) In-context learning, and (4) large language models for clinical tasks.

\subsection{ML with ADRD}
To detect ADRD, \cite{li2023early} employed logistic regression and a gradient boost algorithm using a real-world dataset obtained from the OneFlorida+ Research Consortium. In~\cite{park2020machine}, the use of logistic regression, support vector machines, and random forest methods was explored. The authors of~\cite{nori2019identifying} utilized the LASSO algorithm, and \cite{onishchenko3920640rapid} proposed an algorithm named ZCoR, which assesses the future risk of AD. Differing from the aforementioned studies, which primarily focused on ADRD, the following papers targeted AD more directly~\cite{kumar2021machine}. The authors of~\cite{mohammed2021multi} employed MRI images to detect AD. In~\cite{al2021robust}, the authors utilized EEG signals to detect AD using support vector machines. Several studies have attempted to classify AD using a range of methods, including random forest and linear models~\cite{kleiman2021screening}. K-nearest neighbors (KNN) classifier~\cite{bansal2018comparative}, support vector machine (SVM)~\cite{farid2020applying}, multi-layer perceptron (MLP)~\cite{nagaraj2020risk}, Cuckoo Search~\cite{dhanusha2022robust}, and traditional machine learning techniques~\cite{sanim2022prediction, hurowitz2022dementia}.

\subsection{Large Language Models}
Language models with billions of parameters are commonly referred to as LLMs. The emergence of LLMs began with BERT~\cite{devlin2018bert}, marking the start of the LLM era. GPT-3~\cite{brown2020language} has demonstrated a significant impact on real-world applications due to its versatility. Following GPT-3, GPT-4~\cite{achiam2023gpt} has been developed with improved performance, surpassing human capabilities in some tasks. Very recently, Gemini~\cite{team2023gemini} was announced; however, most models still feature gray-box or black-box architectures, which do not allow access to their model parameters. They only allow to access at most their output probability. To tackle this, Meta release LLaMAv1/v2~\cite{touvron2023llama,touvron2023llama2} which are publicly accessible models for research objectives with various model parameter cases, such as 7B, 13B, and 70B.

\subsection{In-Context Learning}
In-Context Learning is a promising method to make the model answer based on few-shot samples, which dramatically reduces the fine-tuning cost to the target task~\cite{dong2022survey}. The conventional ICL sample selection method randomly selects samples and demonstrates performance improvement~\cite{bolucu2023impact}. To further enhance performance, several works have attempted to select relevant samples using BM25~\cite{robertson2009probabilistic}, and language model-oriented embeddings~\cite{reimers-gurevych-2019-sentence, karpukhin-etal-2020-dense}.

\subsection{LLMs for Clinical Domain}
LLMs possess strong capability in performing various tasks, including those in the medical field~\cite{he2023survey}. In particular, many studies have attempted to develop new LLMs specifically for medical tasks. For example, Med-PaLM~\cite{singhal2023large} represents a medical domain-specific variant of the PaLM model. Similarly, based on Alpaca~\cite{taori2023alpaca}, MedAlpaca~\cite{han2023medalpaca} was proposed, and fine-tuend LLaMA~\cite{touvron2023llama, touvron2023llama2} for medical domain, PMC-LLaMA~\cite{wu2023pmc} was suggested. Chat-bot oriented model~\cite{yunxiang2023chatdoctor} and Huatuo-GPT~\cite{zhang2023huatuogpt} were trained using the dataset obtained from the real-world doctors and ChatGPT~\cite{achiam2023gpt}. Yang et al.~\cite{yang2022gatortron} trained and release the GatorTron model. Different from proposing a new medical-specific models, several works have aimed to directly use the pre-trained LLMs in a zero-shot manner. For example in \cite{nori2023capabilities, liu2023deid} used GPT models for the medical field. Nori et al.~\cite{nori2023can} proposed a way of leveraging pre-trained LLMs for the medical field by leveraging some techniques including in-context learning, and chain-of-thought.
\section{Conclusion}
\label{sec:conclusion}
In this paper, we address the challenge of early ADRD prediction using cost-effective EHR databases. We introduce a novel collaborative approach that combines the predictive power of conventional supervised learning techniques, such as LR, XGB, and MLP, with the advanced reasoning capabilities of LLMs. By integrating a confidence-driven selection process, our method dynamically chooses between the robust, data-driven predictions from SLs and the nuanced, context-aware interpretations by LLMs. This is based on their confidence levels, harnessing the strengths of SLs in clear-cut cases and LLMs in more complex scenarios. The extensive validation of our method with a real-world dataset from OHSU hospital demonstrates the superiority of ADRD risk prediction. Furthermore, our findings reveal that neither scaling up the model size nor specific fine-tuning on medical datasets consistently enhances performance, indicating the need for further investigation into these aspects in practice.

\section*{Acknowledgement}
This material is based in part upon work supported by the National Science Foundation under Grant
IIS-2212174, IIS-1749940, IIS-1750326, Office of Naval Research N00014-24-1-2168, National Institute of General Medical Sciences R01GM134307, R01GM145700, and
National Institute on Aging (NIA) RF1AG072449, R01AG080624, R01AG076448, RF1AG084178, R01AG076234, R01AG080991. 

\bibliographystyle{plain}
\bibliography{ref}

\begin{thebibliography}{10}

\bibitem{achiam2023gpt}
Josh Achiam, Steven Adler, Sandhini Agarwal, Lama Ahmad, Ilge Akkaya, Florencia~Leoni Aleman, Diogo Almeida, Janko Altenschmidt, Sam Altman, Shyamal Anadkat, et~al.
\newblock Gpt-4 technical report.
\newblock {\em arXiv preprint arXiv:2303.08774}, 2023.

\bibitem{al2021robust}
Ali~H Al-Nuaimi, Marina Bl{\=u}ma, Shaymaa~S Al-Juboori, Chima~S Eke, Emmanuel Jammeh, Lingfen Sun, and Emmanuel Ifeachor.
\newblock Robust eeg based biomarkers to detect alzheimer’s disease.
\newblock {\em Brain Sciences}, 11(8):1026, 2021.

\bibitem{albrecht2018predicting}
Jennifer~S Albrecht, Maya Hanna, Dure Kim, and Eleanor~M Perfetto.
\newblock Predicting diagnosis of alzheimer’s disease and related dementias using administrative claims.
\newblock {\em Journal of managed care \& specialty pharmacy}, pages 1138--1145, 2018.

\bibitem{austin2011introduction}
Peter~C Austin.
\newblock An introduction to propensity score methods for reducing the effects of confounding in observational studies.
\newblock {\em Multivariate behavioral research}, 46(3):399--424, 2011.

\bibitem{bansal2018comparative}
Deepika Bansal, Rita Chhikara, Kavita Khanna, and Poonam Gupta.
\newblock Comparative analysis of various machine learning algorithms for detecting dementia.
\newblock {\em Procedia computer science}, 132:1497--1502, 2018.

\bibitem{beam2018big}
Andrew~L Beam and Isaac~S Kohane.
\newblock Big data and machine learning in health care.
\newblock {\em Jama}, 319(13):1317--1318, 2018.

\bibitem{bolucu2023impact}
Necva B{\"o}l{\"u}c{\"u}, Maciej Rybinski, and Stephen Wan.
\newblock Impact of sample selection on in-context learning for entity extraction from scientific writing.
\newblock In {\em Findings of the Association for Computational Linguistics: EMNLP 2023}, pages 5090--5107, 2023.

\bibitem{brown2020language}
Tom Brown, Benjamin Mann, Nick Ryder, Melanie Subbiah, Jared~D Kaplan, Prafulla Dhariwal, Arvind Neelakantan, Pranav Shyam, Girish Sastry, Amanda Askell, et~al.
\newblock Language models are few-shot learners.
\newblock {\em Advances in neural information processing systems}, 33:1877--1901, 2020.

\bibitem{chawla2002smote}
Nitesh~V Chawla, Kevin~W Bowyer, Lawrence~O Hall, and W~Philip Kegelmeyer.
\newblock Smote: synthetic minority over-sampling technique.
\newblock {\em Journal of artificial intelligence research}, 16:321--357, 2002.

\bibitem{chen2015xgboost}
Tianqi Chen, Tong He, Michael Benesty, Vadim Khotilovich, Yuan Tang, Hyunsu Cho, Kailong Chen, Rory Mitchell, Ignacio Cano, Tianyi Zhou, et~al.
\newblock Xgboost: extreme gradient boosting.
\newblock {\em R package version 0.4-2}, 1(4):1--4, 2015.

\bibitem{epfmedtrn}
Zeming Chen, Alejandro Hernández-Cano, Angelika Romanou, Antoine Bonnet, Kyle Matoba, Francesco Salvi, Matteo Pagliardini, Simin Fan, Andreas Köpf, Amirkeivan Mohtashami, Alexandre Sallinen, Alireza Sakhaeirad, Vinitra Swamy, Igor Krawczuk, Deniz Bayazit, Axel Marmet, Syrielle Montariol, Mary-Anne Hartley, Martin Jaggi, and Antoine Bosselut.
\newblock Meditron-70b: Scaling medical pretraining for large language models, 2023.

\bibitem{together2023redpajama}
Together Computer.
\newblock Redpajama: An open source recipe to reproduce llama training dataset, 2023.

\bibitem{devlin2018bert}
Jacob Devlin, Ming-Wei Chang, Kenton Lee, and Kristina Toutanova.
\newblock Bert: Pre-training of deep bidirectional transformers for language understanding.
\newblock {\em arXiv preprint arXiv:1810.04805}, 2018.

\bibitem{dhanusha2022robust}
C~Dhanusha, AV~Senthil~Kumar, and VS~Giridhar~Akula.
\newblock Robust cuckoo search enabled fuzzy neuro symbolic reasoning-based alzheimer’s disease prediction at their earlier stages.
\newblock In {\em Computer Networks and Inventive Communication Technologies: Proceedings of Fifth ICCNCT 2022}, pages 871--886. Springer, 2022.

\bibitem{dong2022survey}
Qingxiu Dong, Lei Li, Damai Dai, Ce~Zheng, Zhiyong Wu, Baobao Chang, Xu~Sun, Jingjing Xu, and Zhifang Sui.
\newblock A survey for in-context learning.
\newblock {\em arXiv preprint arXiv:2301.00234}, 2022.

\bibitem{farid2020applying}
Ahmed~Abdullah Farid, Gamal~Ibrahim Selim, and Hatem Awad~A Khater.
\newblock Applying artificial intelligence techniques to improve clinical diagnosis of alzheimer’s disease.
\newblock {\em European Journal of Engineering Science and Technology}, 3(2):58--79, 2020.

\bibitem{fujiyoshi2017validity}
Akira Fujiyoshi, David~R Jacobs~Jr, Alvaro Alonso, Jos{\'e}~A Luchsinger, Stephen~R Rapp, and Daniel~A Duprez.
\newblock Validity of death certificate and hospital discharge icd codes for dementia diagnosis: the multi ethnic study of atherosclerosis.
\newblock {\em Alzheimer disease and associated disorders}, 31(2):168, 2017.

\bibitem{goldstein2017opportunities}
Benjamin~A Goldstein, Ann~Marie Navar, Michael~J Pencina, and John~PA Ioannidis.
\newblock Opportunities and challenges in developing risk prediction models with electronic health records data: a systematic review.
\newblock {\em Journal of the American Medical Informatics Association: JAMIA}, 24(1):198, 2017.

\bibitem{gong2020tablegpt}
Heng Gong, Yawei Sun, Xiaocheng Feng, Bing Qin, Wei Bi, Xiaojiang Liu, and Ting Liu.
\newblock Tablegpt: Few-shot table-to-text generation with table structure reconstruction and content matching.
\newblock In {\em Proceedings of the 28th International Conference on Computational Linguistics}, pages 1978--1988, 2020.

\bibitem{gruver2024large}
Nate Gruver, Marc Finzi, Shikai Qiu, and Andrew~G Wilson.
\newblock Large language models are zero-shot time series forecasters.
\newblock {\em Advances in Neural Information Processing Systems}, 36, 2024.

\bibitem{han2023medalpaca}
Tianyu Han, Lisa~C Adams, Jens-Michalis Papaioannou, Paul Grundmann, Tom Oberhauser, Alexander L{\"o}ser, Daniel Truhn, and Keno~K Bressem.
\newblock Medalpaca--an open-source collection of medical conversational ai models and training data.
\newblock {\em arXiv preprint arXiv:2304.08247}, 2023.

\bibitem{haykin1998neural}
Simon Haykin.
\newblock {\em Neural networks: a comprehensive foundation}.
\newblock Prentice Hall PTR, 1998.

\bibitem{he2023survey}
Kai He, Rui Mao, Qika Lin, Yucheng Ruan, Xiang Lan, Mengling Feng, and Erik Cambria.
\newblock A survey of large language models for healthcare: from data, technology, and applications to accountability and ethics.
\newblock {\em arXiv preprint arXiv:2310.05694}, 2023.

\bibitem{hegselmann2023tabllm}
Stefan Hegselmann, Alejandro Buendia, Hunter Lang, Monica Agrawal, Xiaoyi Jiang, and David Sontag.
\newblock Tabllm: Few-shot classification of tabular data with large language models.
\newblock In {\em International Conference on Artificial Intelligence and Statistics}, pages 5549--5581. PMLR, 2023.

\bibitem{hermanutz1975clinical}
KD~Hermanutz, A~Wahlen, and A~Sobbe.
\newblock The clinical importance of angiography in the diagnosis of periarteritis nodosa.
\newblock {\em Rontgen-blatter; Zeitschrift fur Rontgen-technik und Medizinisch-wissenschaftliche Photographie}, 28(8):339--348, 1975.

\bibitem{hosmer2013applied}
David~W Hosmer~Jr, Stanley Lemeshow, and Rodney~X Sturdivant.
\newblock {\em Applied logistic regression}, volume 398.
\newblock John Wiley \& Sons, 2013.

\bibitem{hurowitz2022dementia}
Joseph Hurowitz.
\newblock Dementia classification through textual analysis with machine learning algorithms.
\newblock 2022.

\bibitem{jack2010hypothetical}
Clifford~R Jack, David~S Knopman, William~J Jagust, Leslie~M Shaw, Paul~S Aisen, Michael~W Weiner, Ronald~C Petersen, and John~Q Trojanowski.
\newblock Hypothetical model of dynamic biomarkers of the alzheimer's pathological cascade.
\newblock {\em The Lancet Neurology}, 9(1):119--128, 2010.

\bibitem{jin2020disease}
Di~Jin, Eileen Pan, Nassim Oufattole, Wei-Hung Weng, Hanyi Fang, and Peter Szolovits.
\newblock What disease does this patient have? a large-scale open domain question answering dataset from medical exams, 2020.

\bibitem{karpukhin-etal-2020-dense}
Vladimir Karpukhin, Barlas Oguz, Sewon Min, Patrick Lewis, Ledell Wu, Sergey Edunov, Danqi Chen, and Wen-tau Yih.
\newblock Dense passage retrieval for open-domain question answering.
\newblock In Bonnie Webber, Trevor Cohn, Yulan He, and Yang Liu, editors, {\em Proceedings of the 2020 Conference on Empirical Methods in Natural Language Processing (EMNLP)}, pages 6769--6781, Online, November 2020. Association for Computational Linguistics.

\bibitem{kleiman2021screening}
Michael~J Kleiman, Elan Barenholtz, James~E Galvin, Alzheimer’s Disease~Neuroimaging Initiative, et~al.
\newblock Screening for early-stage alzheimer’s disease using optimized feature sets and machine learning.
\newblock {\em Journal of Alzheimer's Disease}, 81(1):355--366, 2021.

\bibitem{kumar2021machine}
Sayantan Kumar, Inez Oh, Suzanne Schindler, Albert~M Lai, Philip~RO Payne, and Aditi Gupta.
\newblock Machine learning for modeling the progression of alzheimer disease dementia using clinical data: a systematic literature review.
\newblock {\em JAMIA open}, 4(3):ooab052, 2021.

\bibitem{kweon2023publicly}
Sunjun Kweon, Junu Kim, Jiyoun Kim, Sujeong Im, Eunbyeol Cho, Seongsu Bae, Jungwoo Oh, Gyubok Lee, Jong~Hak Moon, Seng~Chan You, Seungjin Baek, Chang~Hoon Han, Yoon~Bin Jung, Yohan Jo, and Edward Choi.
\newblock Publicly shareable clinical large language model built on synthetic clinical notes, 2023.

\bibitem{li2023early}
Qian Li, Xi~Yang, Jie Xu, Yi~Guo, Xing He, Hui Hu, Tianchen Lyu, David Marra, Amber Miller, Glenn Smith, et~al.
\newblock Early prediction of alzheimer's disease and related dementias using real-world electronic health records.
\newblock {\em Alzheimer's \& Dementia}, 2023.

\bibitem{li2023finding}
Xiaonan Li and Xipeng Qiu.
\newblock Finding support examples for in-context learning.
\newblock In {\em Findings of the Association for Computational Linguistics: EMNLP 2023}, pages 6219--6235, 2023.

\bibitem{liu2022makes}
Jiachang Liu, Dinghan Shen, Yizhe Zhang, Bill Dolan, Lawrence Carin, and Weizhu Chen.
\newblock What makes good in-context examples for gpt-3?
\newblock {\em DeeLIO 2022}, page 100, 2022.

\bibitem{liu2018table}
Tianyu Liu, Kexiang Wang, Lei Sha, Baobao Chang, and Zhifang Sui.
\newblock Table-to-text generation by structure-aware seq2seq learning.
\newblock In {\em Proceedings of the AAAI conference on artificial intelligence}, volume~32, 2018.

\bibitem{liu2023deid}
Zhengliang Liu, Xiaowei Yu, Lu~Zhang, Zihao Wu, Chao Cao, Haixing Dai, Lin Zhao, Wei Liu, Dinggang Shen, Quanzheng Li, et~al.
\newblock Deid-gpt: Zero-shot medical text de-identification by gpt-4.
\newblock {\em arXiv preprint arXiv:2303.11032}, 2023.

\bibitem{mattson2004pathways}
Mark~P Mattson.
\newblock Pathways towards and away from alzheimer's disease.
\newblock {\em Nature}, 430(7000):631--639, 2004.

\bibitem{mohammed2021multi}
Badiea~Abdulkarem Mohammed, Ebrahim~Mohammed Senan, Taha~H Rassem, Nasrin~M Makbol, Adwan~Alownie Alanazi, Zeyad~Ghaleb Al-Mekhlafi, Tariq~S Almurayziq, and Fuad~A Ghaleb.
\newblock Multi-method analysis of medical records and mri images for early diagnosis of dementia and alzheimer’s disease based on deep learning and hybrid methods.
\newblock {\em Electronics}, 10(22):2860, 2021.

\bibitem{nagaraj2020risk}
Sanjay Nagaraj and Tim~Q Duong.
\newblock Risk score stratification of alzheimer’s disease and mild cognitive impairment using deep learning.
\newblock {\em medRxiv}, pages 2020--11, 2020.

\bibitem{nori2023capabilities}
Harsha Nori, Nicholas King, Scott~Mayer McKinney, Dean Carignan, and Eric Horvitz.
\newblock Capabilities of gpt-4 on medical challenge problems.
\newblock {\em arXiv preprint arXiv:2303.13375}, 2023.

\bibitem{nori2023can}
Harsha Nori, Yin~Tat Lee, Sheng Zhang, Dean Carignan, Richard Edgar, Nicolo Fusi, Nicholas King, Jonathan Larson, Yuanzhi Li, Weishung Liu, et~al.
\newblock Can generalist foundation models outcompete special-purpose tuning? case study in medicine.
\newblock {\em arXiv preprint arXiv:2311.16452}, 2023.

\bibitem{nori2019machine}
Vijay~S Nori, Christopher~A Hane, William~H Crown, Rhoda Au, William~J Burke, Darshak~M Sanghavi, and Paul Bleicher.
\newblock Machine learning models to predict onset of dementia: a label learning approach.
\newblock {\em Alzheimer's \& Dementia: Translational Research \& Clinical Interventions}, 5:918--925, 2019.

\bibitem{nori2019identifying}
Vijay~S Nori, Christopher~A Hane, David~C Martin, Alexander~D Kravetz, and Darshak~M Sanghavi.
\newblock Identifying incident dementia by applying machine learning to a very large administrative claims dataset.
\newblock {\em PLoS One}, 14(7):e0203246, 2019.

\bibitem{onishchenko3920640rapid}
Dmytro Onishchenko, Sam Searle, Kenneth Rockwood, James Mastrianni, and Ishanu Chattopadhyay.
\newblock Rapid universal early screening for alzheimer's disease and related dementia via pattern discovery in diagnostic history.
\newblock {\em Available at SSRN 3920640}.

\bibitem{park2020machine}
Ji~Hwan Park, Han~Eol Cho, Jong~Hun Kim, Melanie~M Wall, Yaakov Stern, Hyunsun Lim, Shinjae Yoo, Hyoung~Seop Kim, and Jiook Cha.
\newblock Machine learning prediction of incidence of alzheimer’s disease using large-scale administrative health data.
\newblock {\em NPJ digital medicine}, 3(1):46, 2020.

\bibitem{pukelsheim1994three}
Friedrich Pukelsheim.
\newblock The three sigma rule.
\newblock {\em The American Statistician}, 48(2):88--91, 1994.

\bibitem{reimers-gurevych-2019-sentence}
Nils Reimers and Iryna Gurevych.
\newblock Sentence-{BERT}: Sentence embeddings using {S}iamese {BERT}-networks.
\newblock In Kentaro Inui, Jing Jiang, Vincent Ng, and Xiaojun Wan, editors, {\em Proceedings of the 2019 Conference on Empirical Methods in Natural Language Processing and the 9th International Joint Conference on Natural Language Processing (EMNLP-IJCNLP)}, pages 3982--3992, Hong Kong, China, November 2019. Association for Computational Linguistics.

\bibitem{robertson2009probabilistic}
Stephen Robertson, Hugo Zaragoza, et~al.
\newblock The probabilistic relevance framework: Bm25 and beyond.
\newblock {\em Foundations and Trends{\textregistered} in Information Retrieval}, 3(4):333--389, 2009.

\bibitem{sanim2022prediction}
Mostofa~Shariar Sanim, Md~Rahatul Islam, Shahid Rahman, Rafi Afzal, and Khan~Mehedi Hasan.
\newblock Prediction of dementia using smote based oversampling and stacking classifier.
\newblock In {\em International Conference on Hybrid Intelligent Systems}, pages 441--452. Springer, 2022.

\bibitem{sha2018order}
Lei Sha, Lili Mou, Tianyu Liu, Pascal Poupart, Sujian Li, Baobao Chang, and Zhifang Sui.
\newblock Order-planning neural text generation from structured data.
\newblock In {\em Proceedings of the AAAI Conference on Artificial Intelligence}, volume~32, 2018.

\bibitem{sharegpt2023sharegpt}
Teams ShareGPT.
\newblock Sharegpt: Share your wildest chatgpt conversations with one click, 2023.

\bibitem{shickel2017deep}
Benjamin Shickel, Patrick~James Tighe, Azra Bihorac, and Parisa Rashidi.
\newblock Deep ehr: a survey of recent advances in deep learning techniques for electronic health record (ehr) analysis.
\newblock {\em IEEE journal of biomedical and health informatics}, 22(5):1589--1604, 2017.

\bibitem{singhal2023large}
Karan Singhal, Shekoofeh Azizi, Tao Tu, S~Sara Mahdavi, Jason Wei, Hyung~Won Chung, Nathan Scales, Ajay Tanwani, Heather Cole-Lewis, Stephen Pfohl, et~al.
\newblock Large language models encode clinical knowledge.
\newblock {\em Nature}, 620(7972):172--180, 2023.

\bibitem{tang2024leveraging}
Alice~S Tang, Katherine~P Rankin, Gabriel Cerono, Silvia Miramontes, Hunter Mills, Jacquelyn Roger, Billy Zeng, Charlotte Nelson, Karthik Soman, Sarah Woldemariam, et~al.
\newblock Leveraging electronic health records and knowledge networks for alzheimer’s disease prediction and sex-specific biological insights.
\newblock {\em Nature Aging}, pages 1--17, 2024.

\bibitem{taori2023alpaca}
Rohan Taori, Ishaan Gulrajani, Tianyi Zhang, Yann Dubois, Xuechen Li, Carlos Guestrin, Percy Liang, and Tatsunori~B Hashimoto.
\newblock Alpaca: A strong, replicable instruction-following model.
\newblock {\em Stanford Center for Research on Foundation Models. https://crfm. stanford. edu/2023/03/13/alpaca. html}, 3(6):7, 2023.

\bibitem{team2023gemini}
Gemini Team, Rohan Anil, Sebastian Borgeaud, Yonghui Wu, Jean-Baptiste Alayrac, Jiahui Yu, Radu Soricut, Johan Schalkwyk, Andrew~M Dai, Anja Hauth, et~al.
\newblock Gemini: a family of highly capable multimodal models.
\newblock {\em arXiv preprint arXiv:2312.11805}, 2023.

\bibitem{tjandra2020cohort}
Donna Tjandra, Raymond~Q Migrino, Bruno Giordani, and Jenna Wiens.
\newblock Cohort discovery and risk stratification for alzheimer's disease: an electronic health record-based approach.
\newblock {\em Alzheimer's \& Dementia: Translational Research \& Clinical Interventions}, 6(1):e12035, 2020.

\bibitem{toma2023clinical}
Augustin Toma, Patrick~R. Lawler, Jimmy Ba, Rahul~G. Krishnan, Barry~B. Rubin, and Bo~Wang.
\newblock Clinical camel: An open expert-level medical language model with dialogue-based knowledge encoding, 2023.

\bibitem{touvron2023llama}
Hugo Touvron, Thibaut Lavril, Gautier Izacard, Xavier Martinet, Marie-Anne Lachaux, Timoth{\'e}e Lacroix, Baptiste Rozi{\`e}re, Naman Goyal, Eric Hambro, Faisal Azhar, et~al.
\newblock Llama: Open and efficient foundation language models.
\newblock {\em arXiv preprint arXiv:2302.13971}, 2023.

\bibitem{touvron2023llama2}
Hugo Touvron, Louis Martin, Kevin Stone, Peter Albert, Amjad Almahairi, Yasmine Babaei, Nikolay Bashlykov, Soumya Batra, Prajjwal Bhargava, Shruti Bhosale, et~al.
\newblock Llama 2: Open foundation and fine-tuned chat models.
\newblock {\em arXiv preprint arXiv:2307.09288}, 2023.

\bibitem{wang2015psf}
Fei Wang and Jimeng Sun.
\newblock Psf: a unified patient similarity evaluation framework through metric learning with weak supervision.
\newblock {\em IEEE journal of biomedical and health informatics}, 19(3):1053--1060, 2015.

\bibitem{wei2023symbol}
Jerry Wei, Le~Hou, Andrew Lampinen, Xiangning Chen, Da~Huang, Yi~Tay, Xinyun Chen, Yifeng Lu, Denny Zhou, Tengyu Ma, et~al.
\newblock Symbol tuning improves in-context learning in language models.
\newblock {\em arXiv preprint arXiv:2305.08298}, 2023.

\bibitem{wei2015extracting}
Wei-Qi Wei and Joshua~C Denny.
\newblock Extracting research-quality phenotypes from electronic health records to support precision medicine.
\newblock {\em Genome medicine}, 7:1--14, 2015.

\bibitem{wilkinson2019identifying}
Tim Wilkinson, Christian Schnier, Kathryn Bush, Kristiina Rannikm{\"a}e, David~E Henshall, Chris Lerpiniere, Naomi~E Allen, Robin Flaig, Tom~C Russ, Deborah Bathgate, et~al.
\newblock Identifying dementia outcomes in uk biobank: a validation study of primary care, hospital admissions and mortality data.
\newblock {\em European journal of epidemiology}, 34:557--565, 2019.

\bibitem{wu2023pmc}
Chaoyi Wu, Xiaoman Zhang, Ya~Zhang, Yanfeng Wang, and Weidi Xie.
\newblock Pmc-llama: Further finetuning llama on medical papers.
\newblock {\em arXiv preprint arXiv:2304.14454}, 2023.

\bibitem{wu2010prediction}
Jionglin Wu, Jason Roy, and Walter~F Stewart.
\newblock Prediction modeling using ehr data: challenges, strategies, and a comparison of machine learning approaches.
\newblock {\em Medical care}, pages S106--S113, 2010.

\bibitem{yang2022gatortron}
Xi~Yang, Aokun Chen, Nima PourNejatian, Hoo~Chang Shin, Kaleb~E Smith, Christopher Parisien, Colin Compas, Cheryl Martin, Mona~G Flores, Ying Zhang, et~al.
\newblock Gatortron: A large clinical language model to unlock patient information from unstructured electronic health records.
\newblock {\em arXiv preprint arXiv:2203.03540}, 2022.

\bibitem{yunxiang2023chatdoctor}
Li~Yunxiang, Li~Zihan, Zhang Kai, Dan Ruilong, and Zhang You.
\newblock Chatdoctor: A medical chat model fine-tuned on llama model using medical domain knowledge.
\newblock {\em arXiv preprint arXiv:2303.14070}, 2023.

\bibitem{zhang2023huatuogpt}
Hongbo Zhang, Junying Chen, Feng Jiang, Fei Yu, Zhihong Chen, Jianquan Li, Guiming Chen, Xiangbo Wu, Zhiyi Zhang, Qingying Xiao, et~al.
\newblock Huatuogpt, towards taming language model to be a doctor.
\newblock {\em arXiv preprint arXiv:2305.15075}, 2023.

\bibitem{zhang2019metapred}
Xi~Sheryl Zhang, Fengyi Tang, Hiroko~H Dodge, Jiayu Zhou, and Fei Wang.
\newblock Metapred: Meta-learning for clinical risk prediction with limited patient electronic health records.
\newblock In {\em Proceedings of the 25th ACM SIGKDD international conference on knowledge discovery \& data mining}, pages 2487--2495, 2019.

\bibitem{zhao2023investigating}
Yilun Zhao, Haowei Zhang, Shengyun Si, Linyong Nan, Xiangru Tang, and Arman Cohan.
\newblock Investigating table-to-text generation capabilities of large language models in real-world information seeking scenarios.
\newblock In {\em Proceedings of the 2023 Conference on Empirical Methods in Natural Language Processing: Industry Track}, pages 160--175, 2023.

\bibitem{zhao2021calibrate}
Zihao Zhao, Eric Wallace, Shi Feng, Dan Klein, and Sameer Singh.
\newblock Calibrate before use: Improving few-shot performance of language models.
\newblock In {\em International Conference on Machine Learning}, pages 12697--12706. PMLR, 2021.

\bibitem{zhou2014micro}
Jiayu Zhou, Fei Wang, Jianying Hu, and Jieping Ye.
\newblock From micro to macro: data driven phenotyping by densification of longitudinal electronic medical records.
\newblock In {\em Proceedings of the 20th ACM SIGKDD international conference on Knowledge discovery and data mining}, pages 135--144, 2014.

\bibitem{zhu2016measuring}
Zihao Zhu, Changchang Yin, Buyue Qian, Yu~Cheng, Jishang Wei, and Fei Wang.
\newblock Measuring patient similarities via a deep architecture with medical concept embedding.
\newblock In {\em 2016 IEEE 16th International Conference on Data Mining (ICDM)}, pages 749--758. IEEE, 2016.

\end{thebibliography}

\appendix
\supptitle

\renewcommand{\thefigure}{S\arabic{figure}} %
\renewcommand{\thealgorithm}{S\arabic{algorithm}} 
\setcounter{figure}{0} %
\setcounter{algorithm}{0} %

\section{Summarization of EHRs to text}
\label{app:summary}
\begin{figure}[h]
    \vspace{2mm}
    \centering
    \includegraphics[width=0.7\columnwidth]{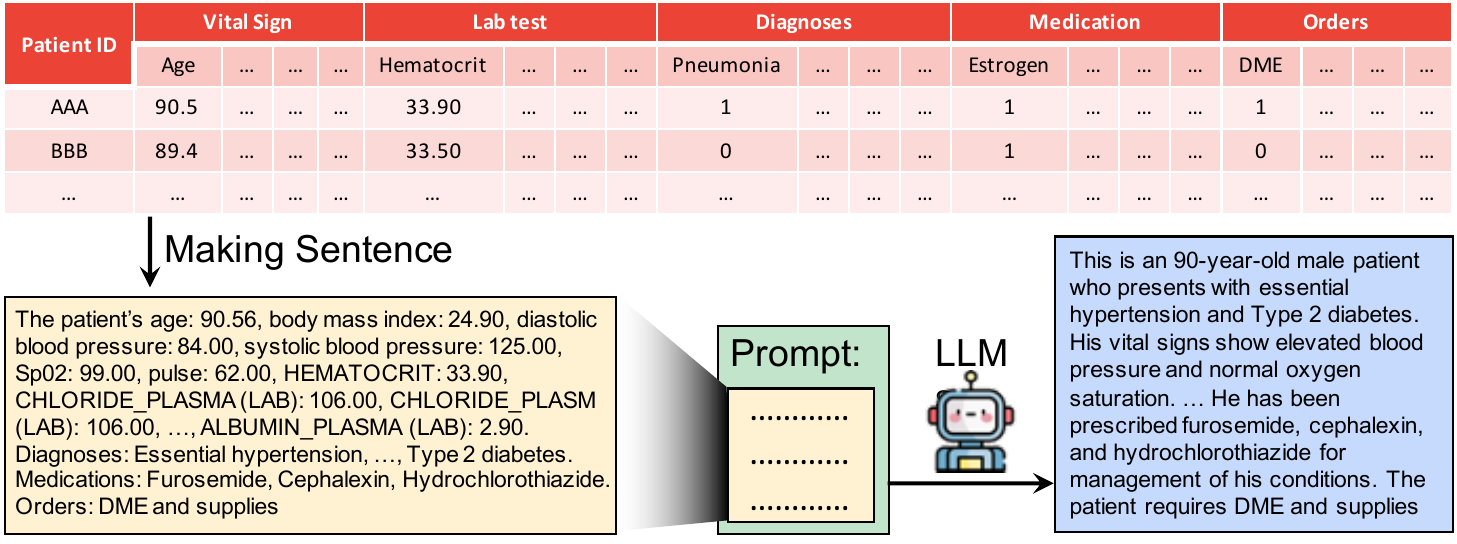}
    \vspace{2mm}
    \caption{Entire pipeline to summarize the tabular format EHR. We first create a sentence format of each patient's information, which is represented in a row, by concatenating them. After that, we leverage LLMs' summarization capability by feeding the concatenated sentence with a guiding prompt for the summarization task.}
    \label{fig:summarization}
\end{figure}

In our proposed pipeline, the EHR data used is tabular data, which cannot be directly utilized by LLM and needs to be further converted to text. For this reason, we develop a method to convert each row in the tabular dataset into an LLM-friendly summary of patient information. Various serialization methods have been explored for this purpose, including text templates, table-to-text models~\cite{liu2018table,sha2018order}, and LLMs~\cite{gong2020tablegpt,zhao2023investigating}. Our method utilizes the LLM to generate patient summaries from tabular data, leveraging its extensive pre-trained knowledge to produce summaries without the need for labeled training data. These LLM-generated summaries closely align with the training distribution of LLM, thereby aiding in subsequent tasks~\cite{hegselmann2023tabllm}. Our proposed pipeline for summarizing EHR into text is shown in~\autoref{fig:summarization}.
\subsection{Concatenating tabular format.} 
\label{sup.concatenating}
Initially, we transform the tabular data of $i^{\text{th}}$ sample (\textcolor{red}{red table} in~\autoref{fig:summarization}) into concatenated sentences $S_i^\text{cat}$ (\textcolor{goldenrod}{yellow box} in~\autoref{fig:summarization}) using following template:
\begin{shaded}
\noindent
\underline{\textbf{Format template $\mc{T}_{\text{cat}}(\mathbf{x}_i)$}} \\
\\
{The patient's \{$f_1$\}: $x_1$, \{$f_2$\}: $x_2$, $\cdots$. Diagnoses: [ICD]. Medications: [RxNorm]. Orders: [CPT].}\\~\\
{\textbf{Output:} $S_i^{\text{cat}}$}
\end{shaded}
\noindent
Here, \{$\cdot$\} is used to denote the description of features. For example, if $x_a$ is the feature value of \code{Age}, then \{$f_a$\} would correspond to the term ``age.'' The operator [$\cdot$] denotes the concatenation of descriptions of active features within categories such as Diagnoses, Medications, and Orders. For example, consider a patient identified as \code{AAA} who has $x_j = 1$ and $x_k = 1$—indicating the presence of the respective conditions—for features $x_j$ and $x_k$, corresponding to \code{Essential hypertension} and \code{Diabetes Type 2}. In this case, the concatenated feature description [ICD] would be ``Essential hypertension, Type 2 diabetes,'' representing the active health conditions of the patient as specified by their respective feature indicators.

\subsection{LLM-based summarization.} 
After obtaining the concatenated  sentences from the raw tabular data using a format template, we summarized them using the LLM, \ie
\begin{equation*}
    S_i = \code{LLM}\left (\mc{T}_{\text{summary}}(S_{i}^{\text{cat}})\right ),
\end{equation*}
where the guideline prompt $\mc{T}_{\text{summary}}$ is defined as follows:
\begin{shaded}
\noindent
\underline{\textbf{LLM's prompt $\mc{T}_{\text{summary}}(S_{i}^{\text{cat}})$}} \\
\\
{You are an expert medical professional. Please summarize the patient's medical record in one paragraph.} \\
{Record: \underline{$S_i^{\text{cat}}$}} \\
{Summary: } \\~\\
{\textbf{Output:} $S_i$}
\end{shaded}
\noindent
In this prompt, the input $S_i^{\text{cat}}$ represents the concatenated sentence for $i^{\text{th}}$ patient as mentioned in \textcolor{ForestGreen}{section}~\ref{sup.concatenating}, the ouput $S_i$ is the patient's summary given by the LLM. In the \textcolor{blue}{blue box} of~\autoref{fig:summarization}, we present an example of the summarized sentence, which is more concise and free from the excessive numbers that LLMs typically struggle with~\cite{gruver2024large}. In our use of the LLaMA2 7B model for EHR summarization, we observed that the ``repetition\_penalty'' parameter significantly affects the quality of the summaries produced. When set too low ($1.0$), the summaries tend to omit explanations of certain medical terms. Conversely, when set too high ($1.5$), the summaries occasionally generate fictitious content, or what could be described as hallucinations. We recommend using a ``repetition\_penalty'' setting of either $1.1$ or $1.2$.
\section{Experimental Setting of the Baselines}
\subsection{Model Training and Validation}

We employ a unified framework for training different SLs, including LR, XGB, and MLP, which are served as baseline classifiers in our experiment. These models are selected for their scalability and their ability to manage different type of feature relationships (from linear to non-linear) in medical tabular data. Each model is tuned using a 5-fold cross-validation approach, facilitated by the \texttt{StratifiedKFold} from scikit-learn, configured with \texttt{n\_splits=5}, \texttt{shuffle=True}, and \texttt{random\_state=42}. This step ensures that the optimal hyperparameters obtained through cross-validation reflect the overall preferences of the dataset.

\subsection{Hyperparameter Tuning}

We conduct hyperparameter tuning using \texttt{GridSearchCV} for LR and MLP, and \texttt{RandomizedSearchCV} for XGB, aiming to optimize the F1 score. The specific hyperparameter grids for each model are as follows:

\vspace{1em}
\noindent
\emph{Logistic Regression (LR).}

\begin{table}[h]
    \centering
    \resizebox{0.9\textwidth}{!}{
    \begin{tabular}{ll}
    \thickhline
        Parameter & Values \\ \hline
        \code{solver} & Options include [\code{sag}, \code{saga}, \code{lbfgs}] \\
        \code{C} & Regularizaation values [1e-4, 1e-3, 1e-2, 1e-1, 1, 1e1, 1e2, 1e,3, 1e4\\
        \code{tol}  & Tolerance for stopping criateria [1e-4, 5e-4, 1e-3, 5e-3, 1e-2, 5e-2, 1e-1, 5e-1] \\
        \thickhline
    \end{tabular}}

\end{table}

\noindent
\emph{XGBoost (XGB).}

\begin{table}[h]
    \centering
    \resizebox{0.9\textwidth}{!}{
    \begin{tabular}{ll}
    \thickhline
        Parameter & Values \\ \hline
        \code{num\_leaves} & Range from 2 to 256, stepping by 4 \\
        \code{learning\_rate} & Values [0.1, 0.05, 0.01, 0.005, 0.001, 0.0005, 0.0001, 0.00005, 0.00001] \\
        \code{n\_estimators} & Range from 100 to 2000, stepping by 100 \\
        \code{min\_split\_gain} & Values [1e-3, 1e-4, 1e-5, 1e-6, 1e-7]\\ 
        \code{min\_child\_weight} & Values [1e-5, 1e-3, 1e-2, 1e-1, 1, 1e1] \\
        \code{min\_child\_samples} & Options include [8, 16, 32, 48, 64, 80, 96, 128, 256, 396] \\
        \code{subsample}, \code{colsample\_bytree} & Ranges from 0.2 to 1, stepping by 0.1 \\
        \code{reg\_alpha}, \code{reg\_lambda} & Values [0.25, 0.5, 0.75, 1, 2, 5, 10] \\
        \code{tree\_learner} & Options include [\code{feature}, \code{data}, \code{voting}] \\
        \code{max\_bin} & Options include [128, 256, 512] \\
        \code{boosting\_type} & Types [\code{dart}, \code{gbdt}]\\
        \thickhline
    \end{tabular}}

\end{table}

\vspace{3em}
\noindent
\emph{Multi-layer Perceptron (MLP).}

\begin{table}[h]
    \centering
    \resizebox{0.9\textwidth}{!}{
    \begin{tabular}{ll}
    \thickhline
        Parameter & Values \\ \hline
        \code{hidden\_layer\_sizes} &  Configurations [(50,), (100,), (50, 50), (50, 100), (100, 100), (50, 50, 50)]\\
        \code{activation} &  Options [\code{tanh}, \code{relu}]\\
        \code{solver} &  Options [\code{sgd}, \code{adam}] \\
        \code{alpha} &  Values [1e-5, 5e-5, 1e-4, 5e-4, 1e-3, 5e-3, 1e-2, 5e-2, 1e-1, 5e-1] \\
        \code{learning\_rate} &  Modes [\code{constant}, \code{adaptive}] \\
        \thickhline
    \end{tabular}}

\end{table}

\section{Obtaining training samples' prediction probability}
In order to obtain the prediction probabilities for each sample in the training dataset $D_{\text{train}} = \{(\mathbf{x}_i, y_i)\}_{i=1}^{n}$, we do not train the model directly on the entire training dataset. Instead, the $D_{\text{train}}$ is further partitioned into training and testing folds, with this partitioning process being repeated $k=10$ times in order to ensure that each sample in $D_{\text{train}}$ is covered by a testing fold. The model is then trained on the training fold and subsequently used to predict the probabilities for each sample in the testing fold. The algorithm is as follows:

\begin{algorithm}
    \caption{Obtaining Training samples' Prediction Probabilities }
    \begin{algorithmic}
        \State \textbf{Input: } $k$, $D_{\text{train}}$
        \vspace{2mm}
        \State Partition $D_{\text{train}}$ into $k$ equal folds
        \For{$i = 1$ to $k$}
            \State Designate $i$-th fold as the test set, and the remaining as the training set
            \State Train the model on the training set
            \State Predict the probabilities on the test set
            \State Store the prediction probabilities for samples in $i$-th fold
        \EndFor
        \vspace{2mm}
        \State \textbf{Output: } Aggregated prediction probabilities from all folds
    \end{algorithmic}
\end{algorithm}

This method provides a fair assessment of the model’s predictive capabilities and ensures that the prediction probabilities are both reliable and representative of the model's actual performance, which helps avoid overfitting to the entire training set.
\section{Medical LLMs Introduciton}
In our empirical analysis, we explore the performance of LLMs across different model sizes, as well as the performance after being fine-tuned on specific medical data. These medical LLMs are all fine-tuned based on the LLaMA2 model. We obtain the checking-points from Huggingface. The details are as follows:
\footnotetext[1]{\url{https://huggingface.co/wanglab/ClinicalCamel-70B}}
\footnotetext[2]{\url{https://huggingface.co/augtoma/qCammel-70-x}}
\footnotetext[3]{\url{https://huggingface.co/epfl-llm/meditron-70b}}
\footnotetext[4]{\url{https://huggingface.co/starmpcc}}
\footnotetext[5]{\url{https://www.ncbi.nlm.nih.gov/pmc/}}
\begin{itemize}
  \item \textbf{ClinicalCamel-70b}\textsuperscript{1}: \ Clinical Camel~\cite{toma2023clinical} is fine-tuned on the LLaMA2 70B model using QLoRA. It is tailored for the medical and clinical research, capable of processing and generating relevant content. The data used in fine-tuning: (1) 100,000 synthetic dialogues produced via dialogue-based knowledge encoding (DBKE). (2) 10,187 USMLE questions~\cite{jin2020disease} which were converted via DBKE. (3) 70,000 multi-step conversations from the ShareGPT~\cite{sharegpt2023sharegpt} dataset.
  \item \textbf{qCammel-70b}\textsuperscript{2}: \  qCammel is a fine-tuned version of LLaMA2 70B model, trained on a distilled dataset of 15,000 instructions using QLoRA. This model is optimized for academic medical knowledge and instruction-following capabilities.
  \item \textbf{Meditron-70b}\textsuperscript{3}: \ Meditron is a 70B model adapted to the medical domain from LLaMA2 70B model through fine-tuning on a comprehensively curated medical corpus, including selected PubMed articles, abstracts, a dataset of internationally-recognized medical guidelines~\cite{epfmedtrn}, and general domain data from RedPajama-v1~\cite{together2023redpajama}. 
  \item \textbf{Asclepius-7b, 13b}\textsuperscript{4}: \ Asclepius~\cite{kweon2023publicly} is a language model designed for various clinical NLP tasks. It was fine-tuned on the LLaMA2 model, using a dataset of 158,000 high-quality synthetic clinical notes derived from anonymized case reports on PubMed Central\textsuperscript{5}.
  \item \textbf{Asclepius-7b, 13b (w/o instruction tuning)}\textsuperscript{4}:\ This is the variant of Asclepius, without using the instruction tuning.
\end{itemize}

\end{document}